\documentclass[sigconf]{acmart}
\usepackage[utf8]{inputenc}
\usepackage{textgreek}
\usepackage{subcaption}

\AtBeginDocument{%
  }

\copyrightyear{2026}
\acmYear{2026}
\setcopyright{cc}
\setcctype{by-nc-nd}
\acmConference[MM '26]{Proceedings of the 34th ACM International Conference on Multimedia}{November 10--14, 2026}{Rio de Janeiro, Brazil}
\acmBooktitle{Proceedings of the 34th ACM International Conference on Multimedia (MM '26), November 10--14, 2026, Rio de Janeiro, Brazil}
\acmDOI{10.1145/3767308.3836311}
\acmISBN{979-8-4007-2213-4/2026/11}

\usepackage{multirow}
\usepackage{booktabs}
\usepackage{xcolor}
\usepackage{colortbl}
\usepackage{siunitx}

\newcommand{\best}[1]{\textbf{#1}}
\newcommand{\second}[1]{\underline{#1}}

\begin{document}

\title[PAR: Perturb, Aggregate then Reconstruct for Multi-View Anomaly Detection]
{Perturb, Aggregate then Reconstruct: Consensus Aggregation and Fourier-Attended Multi-Decoders for Reconstruction-based
Multi-View Anomaly Detection}

\author{Yongho Son}
\affiliation{%
  \institution{Hanyang University}
  \city{Seoul}
  \country{Republic of Korea}}
\email{yhson@hanyang.ac.kr}
\orcid{0009-0009-4114-1970}

\author{Junyeong Yun}
\affiliation{%
  \institution{Hanyang University}
  \city{Seoul}
  \country{Republic of Korea}}
\email{yuns0713@hanyang.ac.kr}
\orcid{0009-0001-4883-6558}

\author{Je Hyeong Hong}
\authornote{Corresponding author}
\affiliation{%
  \institution{Hanyang University}
  \city{Seoul}
  \country{Republic of Korea}
}\email{jhh37@hanyang.ac.kr}
\orcid{0000-0003-2797-553X}

\renewcommand{\shortauthors}{Yongho Son, Junyeong Yun, and Je Hyeong Hong}

\begin{abstract}
Multi-view anomaly detection is crucial for industrial inspection to identify defects hidden from single viewpoints. Yet, despite their strong success in single-view applications, reconstruction-based methods remain largely underexplored in multi-view settings. A key challenge in extending these methods lies in generating view-consistent pseudo-anomalies. To address this, we propose PAR, a reconstruction-based framework for enhanced multi-view anomaly detection. First, we introduce a feature perturbation strategy coupled with a consensus aggregation module that integrates these perturbed features across viewpoints to mitigate view-inconsistent anomalies and construct highly aligned pseudo-anomaly representations. Second, we employ a view-specialized multi-decoder architecture; by assigning a dedicated decoder to each view, we increase the model's representational complexity, enabling the generation of highly feasible, localized anomaly maps rather than less precise, view-agnostic reconstructions. Last, we incorporate Fourier attention into the decoders to emphasize global frequency-domain structures over local token evidence, effectively preventing identity mapping. Extensive experiments on the Real-IAD dataset demonstrate that PAR consistently outperforms state-of-the-art methods in sample-level, image-level, and pixel-level anomaly detection. Our code is available at \href{https://github.com/SpatialAILab/PAR_mvad}{https://github.com/SpatialAILab/PAR\_mvad}.

\end{abstract}






\maketitle

\begin{figure}[t]
    \centering
    \begin{subfigure}{\linewidth}
        \centering
        \includegraphics[width=\linewidth]{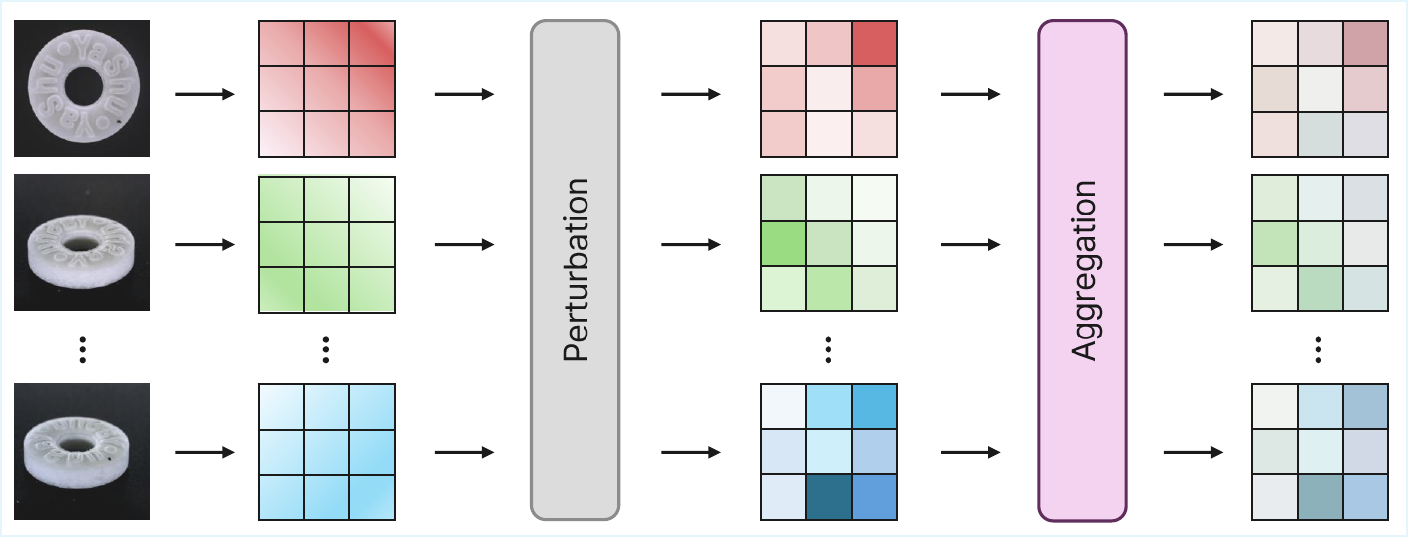}
        %\Description={Multi-view pseudo anomaly generation module}
        \caption{Proposed perturbation with consensus aggregation strategy for producing more aligned view-consistent pseudo-anomalies}
        \label{fig:teaser_a}
    \end{subfigure}

    \vspace{0.5em}

    \begin{subfigure}{\linewidth}
        \centering
        \includegraphics[width=\linewidth]{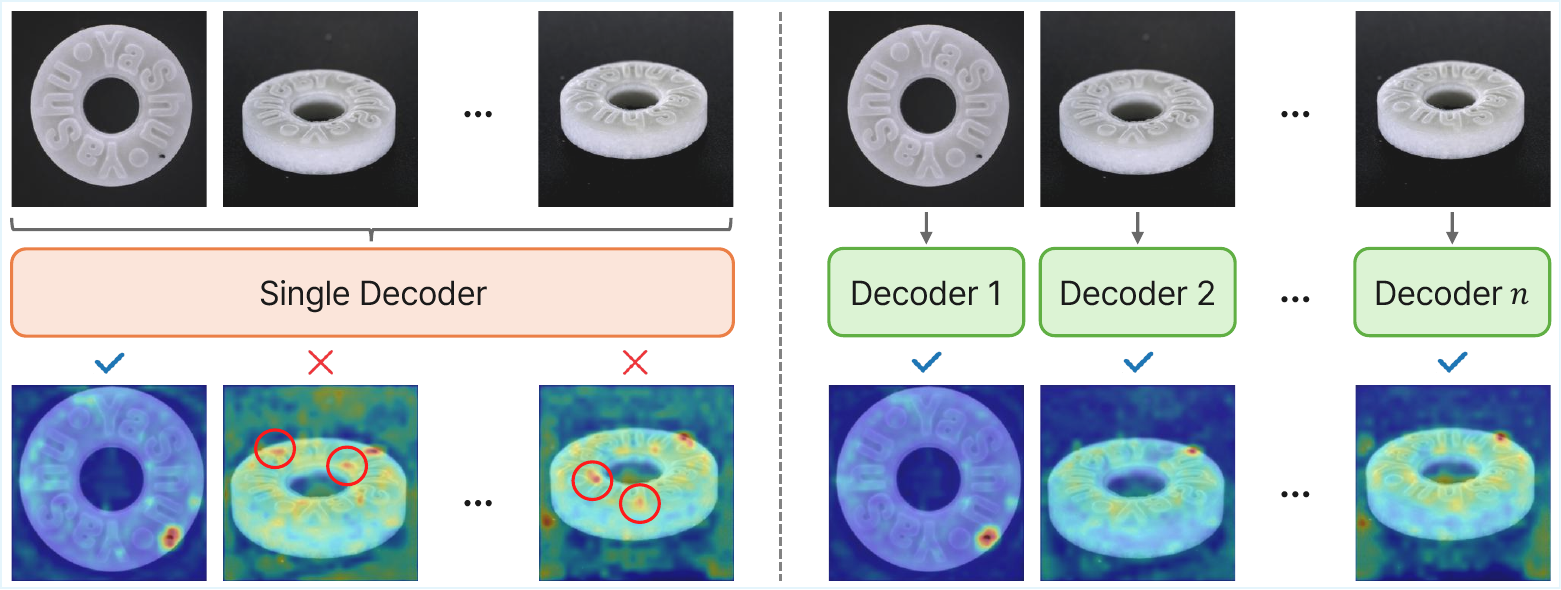}
        %\Description={Multi-decoder architecture}
        \caption{Proposed view-specialized multi-decoder architecture}
        \label{fig:teaser_b}
    \end{subfigure}

\caption{
Overview of the key components of PAR.
(a) Given multi-view inputs, feature perturbation first injects anomaly corruption.
The consensus aggregation module then integrates these complementary cues across viewpoints to construct multi-view-consistent pseudo-anomaly features.
(b) A shared decoder can produce view-agnostic reconstructions and inaccurate localization (red circles), whereas view-specialized decoders enable more accurate and consistent anomaly localization across viewpoints.
}
    \label{fig:teaser}
\end{figure}

\section{Introduction}

Anomaly detection plays a crucial role in industrial inspection, where early defect discovery can reduce material waste, manual inspection costs, and production risk. Since anomalous samples in manufacturing are scarce and diverse, enumerating all possible defect patterns is impractical. Therefore, most visual anomaly detection methods adopt a one-class setting, training only on normal data to detect deviations at test time.

Despite their success, most conventional anomaly detection methods assume single-view images. This is restrictive in real-world inspection, as defect visibility heavily depends on object geometry and viewpoint. While defects on relatively flat objects (e.g., the \textit{PCB} category in Real-IAD, Figure~\ref{fig:qualitative_results}) may remain observable across multiple angles, defective regions on volumetric objects can easily become invisible from certain viewpoints. For example, a bird's-eye view of a \textit{mint} (Figure~\ref{fig:teaser}) cannot reliably capture side-region defects visible only obliquely. Moreover, multi-view anomaly detection must not only observe defects from multiple angles, but also reason about cross-view correlations and geometric inconsistencies. These limitations have sparked recent research in multi-view anomaly detection \cite{wang2024realiad}.

Recent multi-view studies have mainly explored memory bank, teacher-student, and normalizing flow paradigms. For instance, MVAD uses a teacher-student framework for cross-view interactions \cite{he2026learning}, while Multi-Flow enables cross-view message passing within normalizing flows \cite{kruse2025multiflow}. Nevertheless, to the best of our knowledge, no prior industrial multi-view method has systematically explored a reconstruction-based design. Given their strong performance in single-view settings (e.g., RealNet~\cite{zhang2024realnet}, Dinomaly~\cite{guo2025dinomaly}), extending reconstruction-based approaches to multi-view settings provides the primary motivation for our work.

Devising a reconstruction-based multi-view approach, however, turns out to be non-trivial. A key bottleneck is generating view-consistent pseudo-anomalies during training. While local perturbations effectively prevent trivial identity mapping in single-view settings, applying them independently across multiple views fails to incorporate underlying geometric correlations, generating unaligned, isolated corruptions. This  produces overly view-specific features that fail to reflect a coherent, shared anomalous pattern, significantly degrading multi-view anomaly detection performance.

In this work, we address this challenge by proposing PAR (Perturb, Aggregate then Reconstruct), a novel reconstruction-based framework tailored for multi-view anomaly detection. Our pipeline begins by subjecting encoded multi-view features to a stochastic perturbation strategy, followed immediately by a late consensus aggregation module. This step acts as a latent-space consensus mechanism, directly mitigating spatial disjointedness to construct view-aligned pseudo-anomalies. These aligned representations are then routed into a view-specialized multi-decoder architecture. Rather than forcing a single shared decoder to generalize across all camera angles, assigning a dedicated reconstruction pathway to each view substantially increases the model's representational complexity, yielding highly feasible, localized anomaly maps. Furthermore, we equip these decoders with Fourier attention to shift the reconstruction process from the spatial to the frequency domain. By forcing the network to recover global structural patterns instead of copying local token evidence, this mechanism prevents trivial identity mapping and ensures robust, anomaly-free recovery.

In summary, our main contributions are as follows:
\begin{itemize}

    \item We propose \textbf{PAR}, a reconstruction-based framework for multi-view anomaly detection that follows the principle of perturb, aggregate, then reconstruct.
    
    \item We introduce a \emph{composite feature perturbation strategy} combined with \emph{consensus aggregation} to construct better view-aligned pseudo-anomaly features.
    
    \item We design a \emph{view-specialized multi-decoder} architecture equipped with \emph{Fourier attention} to reconstruct features based on global structural patterns, yielding more precise anomaly maps that lead to enhanced multi-view anomaly detection.
\end{itemize}

\begin{figure*}
\centering
\includegraphics[width=0.8\textwidth]{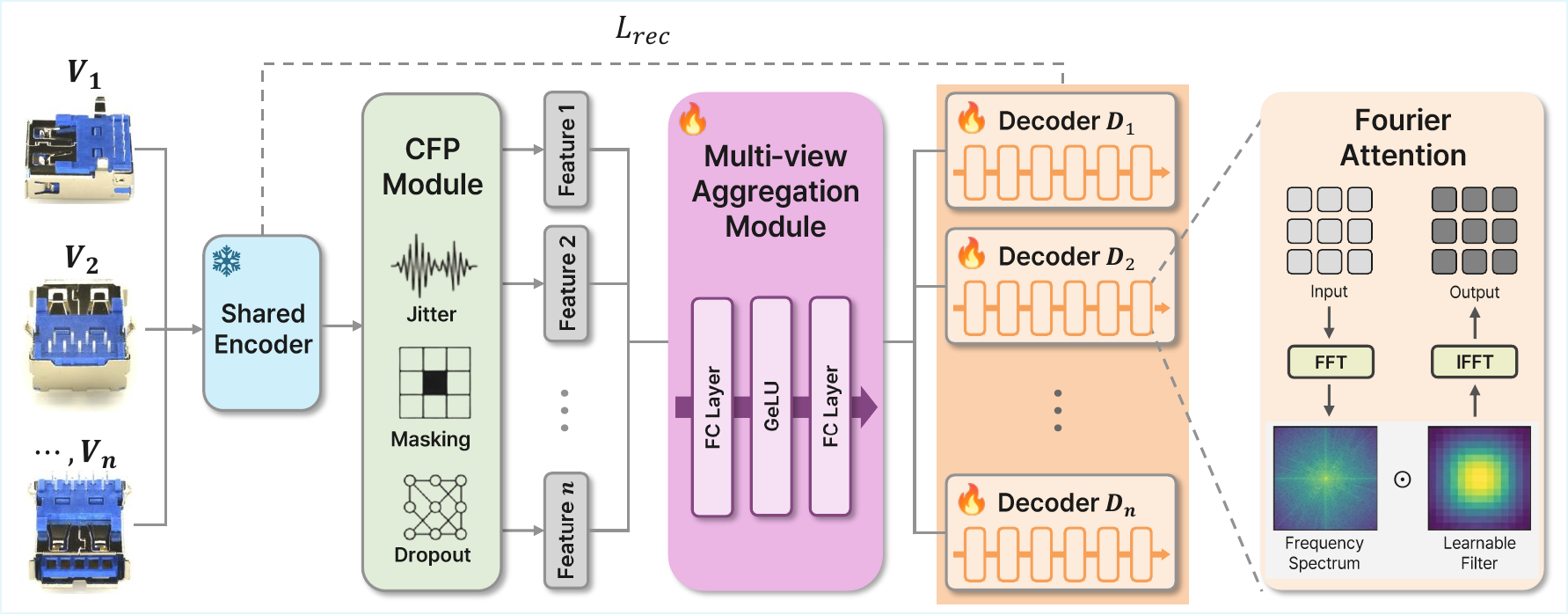}
\caption{Overview of the proposed framework for multi-view anomaly detection. Multi-view images are first encoded by a pre-trained encoder. The resulting features are then perturbed by the composite feature perturbation (CFP) module using feature jitter, masking, and dropout to generate anomaly-like corrupted features. These perturbed features are aggregated across views by the consensus aggregation module to construct more view-aligned pseudo-anomaly features. Finally, the aggregated features are fed into a set of view-specialized decoders, where each decoder is responsible for reconstructing the normal feature of a particular view. Fourier attention is employed in each decoder to emphasize global structural patterns during reconstruction. The model is trained with the reconstruction loss $\mathcal{L}_{rec}$.}
\label{fig:architecture}
\end{figure*}

\section{Related work}

\subsection{Unsupervised anomaly detection}

Unsupervised anomaly detection aims to identify anomalies using only normal data, making it particularly suitable for real-world industrial scenarios where anomalous samples are scarce and diverse. Since it is impractical to collect and annotate all possible defect types, learning solely from normal data provides a scalable and cost-effective solution.
Existing methods can be broadly categorized into representation-based and reconstruction-based approaches. 

Representation-based methods model the distribution of normal features in a representation space and detect anomalies by measuring feature deviations between normal and anomalous samples. PaDiM \cite{defard2021padim} models patch-wise feature distributions with multivariate Gaussians, PatchCore \cite{roth2022patchcore} constructs a memory bank of normal patch features for nearest-neighbor retrieval, and CFA \cite{lee2022cfa} improves localization by adapting target-oriented features with a scalable memory bank. Recent studies have further extended embedding-based approaches using normalizing flow frameworks, such as CFlow \cite{gudovskiy2022cflow}, which explicitly model the probability distribution of normal features to estimate anomaly likelihoods. In addition, recent representation-based approaches have explored diverse feature modeling and embedding strategies, including contrastive learning, prototypical learning, and hybrid reconstruction–embedding frameworks \cite{li2021cutpaste,guo2023recontrast,liu2023simplenet,you2022unified}. However, these methods are often sensitive to distribution shifts and may struggle to generalize under varying conditions. Although these methods achieve strong performance, they also predominantly assume single-view inputs and do not explicitly exploit complementary evidence distributed across multiple viewpoints.

Reconstruction-based methods learn to restore normal images or features and detect anomalies based on reconstruction discrepancies.
DRAEM \cite{zavrtanik2021draem} and NSA \cite{schlueter2022nsa} improve anomaly learning by introducing synthetic anomalies during training, while RealNet \cite{zhang2024realnet} further advances feature reconstruction with realistic synthetic anomalies and adaptive feature selection.
Other reconstruction-oriented methods~\cite{bergmann2020uninformed,wang2021student}, including RD4AD \cite{deng2022rd4ad}, RD++ \cite{tien2023rdpp}, and DeSTSeg \cite{zhang2023destseg}, reconstruct normal feature representations through reverse distillation and feature alignment between teacher and student networks.
Adversarial learning-based approaches have been introduced to improve reconstruction fidelity and better capture complex data distributions \cite{schlegl2019f,akcay2018ganomaly,akccay2019skip}.
Memory-augmented reconstruction methods further constrain the model to represent normal patterns using a limited memory bank, alleviating over-generalization \cite{gong2019memorizing,park2020learning}.
More recently, diffusion-based generative models have been explored to achieve high-quality and robust reconstruction under complex data distributions \cite{zhang2023unsupervised,he2024diffusion}.
Despite these advances, reconstruction-based methods inherently suffer from a normality bias, where both normal and anomalous inputs are projected onto the learned normal manifold, often leading to limited anomaly sensitivity and degraded localization performance.

As a result, existing unsupervised anomaly detection methods are not well aligned with practical multi-view inspection settings, where complementary observations from different viewpoints must be jointly leveraged for reliable anomaly detection and localization.

\subsection{Multi-view anomaly detection}
Multi-view learning has been extensively studied in the broader computer vision community, particularly for tasks such as 3D reconstruction, object recognition, and cross-view representation learning, where complementary information from multiple viewpoints is leveraged to improve robustness and performance \cite{su2015multi,chen2017multi,wang2021patchmatchnet,wang2022mvster,gordon2022flex,zhou2022mvsalnet,pang2023caver,ma2022ppt}. Building upon this foundation, multi-view anomaly detection extends conventional anomaly detection by exploiting multiple observations of the same object to enhance detection reliability and robustness. Unlike single-view approaches, where anomalies may be occluded or only partially visible, multi-view settings enable the aggregation of complementary evidence across views, thereby facilitating more accurate anomaly detection and localization.

Recent studies have explored various strategies to incorporate multi-view information into anomaly detection frameworks. MVAD \cite{he2026learning} proposes an adaptive selection-based attention mechanism that models semantic correlations across views by selecting top-k relevant regions for feature fusion. Multi-Flow \cite{kruse2025multiflow} introduces a normalizing flow-based framework that enables cross-view information exchange through message passing and likelihood estimation across views.
In addition, alignment-based approaches have been investigated to enforce consistency across views. Epipolar-based methods \cite{liu2025multi} utilize geometric constraints to establish correspondences along epipolar lines, while homography-based methods \cite{chen2026unsupervised} perform explicit feature alignment by projecting corresponding regions across views. These approaches aim to maintain spatial consistency and improve cross-view correspondence in anomaly detection tasks.

\noindent\textbf{Remark.}
Despite these advances, effectively leveraging multi-view information for anomaly detection remains challenging. In particular, balancing cross-view interaction and view-specific representations entails an inherent trade-off: excessive information sharing weakens view-specific discriminative cues, while limited interaction hinders the integration of complementary multi-view evidence. Furthermore, while memory banks and normalizing flows have been adapted for multi-view settings, reconstruction-based paradigms remain largely underexplored. We conjecture this is primarily due to the difficulty of generating view-consistent pseudo-anomalies without disrupting the underlying geometric correlations across cameras. Designing an efficient framework that can simultaneously promote cross-view anomaly consistency while preserving view-aware structural reconstruction remains an open problem, which this work seeks to address.

\section{Proposed method}
We now illustrate \emph{Perturb, Aggregate then Reconstruct (PAR)}, a simple yet effective reconstruction-based framework for unsupervised multi-view industrial anomaly detection (see Figure~\ref{fig:architecture} for an overview).

\noindent\textbf{Preliminaries and overview.} \quad
We define the training set as
$\mathcal{X}:=\{\mathbf{I}_i\}_{i=1}^{N}$, where
$\mathbf{I}_i=\{I_i^{(v)}\}_{v=1}^{V}$ denotes the $i$-th multi-view
sample, $I_i^{(v)}\in\mathbb{R}^{H\times W\times 3}$ is the image
captured from the $v$-th view, $H$ and $W$ denote its height and width,
respectively, $V$ is the number of views, and $N$ is the number of
training samples. Given a normal multi-view sample $\mathbf{I}_i$, a shared pre-trained
encoder $\mathcal{E}$ extracts view-wise patch features
$\{F_i^{(v)}\}_{v=1}^{V}$. PAR first perturbs these features using
$\mathcal{P}$ and then integrates the resulting representations through the
consensus aggregation module $\mathcal{A}$. The aggregated features are
subsequently reconstructed by a set of view-specialized decoders
$\{\mathcal{D}^{(v)}\}_{v=1}^{V}$ equipped with Fourier attention, as
illustrated in Figure~\ref{fig:architecture}.

\begin{figure}
\centering
\includegraphics[width=1.0\columnwidth]{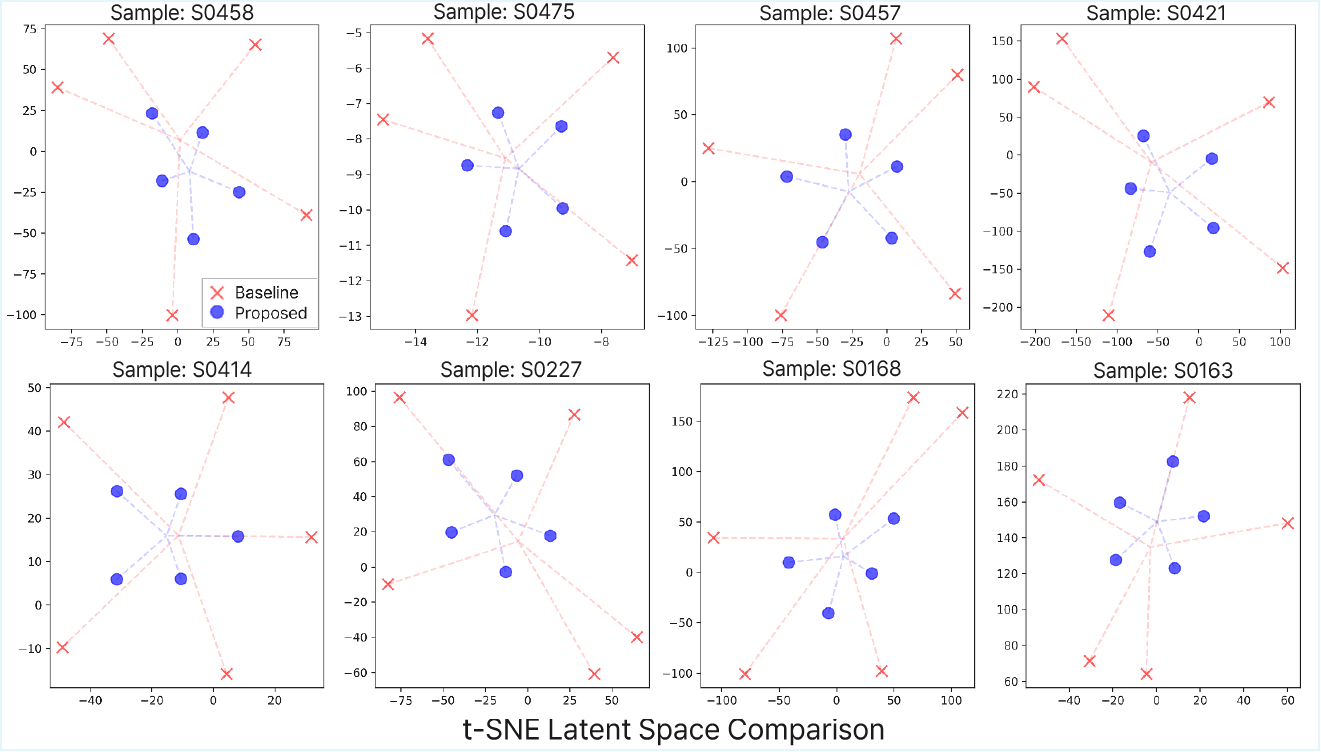}
%\Description={t-SNE visualization of latent pseudo-anomaly features}
\caption{
A $t$-SNE visualization of latent pseudo-anomaly features generated by the vanilla individual-view perturbation strategy (red \textit{x} markers) and our consensus aggregation method (blue dots). More compact feature distributions indicate pseudo-anomaly features are better aligned across views.
}
\label{fig:tsne_feature_vis}
\end{figure}

\subsection{Perturbation and consensus aggregation}
\label{sec:pseudo_anomaly}

A key challenge in extending reconstruction-based anomaly detection to the multi-view setting lies in generating view-aligned pseudo-anomalies. In single-view applications, local perturbations effectively force the model to learn normal structural restoration from corrupted inputs. However, directly applying this strategy to multi-view data is non-trivial. Because each view observes the object from a different angle, independent perturbations often introduce strong view-specific biases, resulting in isolated corruptions rather than a spatially coherent anomalous pattern. To address this, we introduce a consensus aggregation strategy: we first perturb the features of each view independently, and then aggregate them to construct highly aligned and cross-view coherent pseudo-anomaly representations.

Given a normal multi-view sample $\mathbf{I}_i=\{I_i^{(v)}\}_{v=1}^{V}$, we first extract patch-level features from each view using a shared pre-trained encoder $\mathcal{E}$. The encoded feature of the $v$-th view is denoted as
\begin{equation}
F_i^{(v)}
=
\mathcal{E}\!\left(I_i^{(v)}\right)
=
\left[
f_{i,1}^{(v)},\ldots,f_{i,J}^{(v)}
\right]^{\top}
\in\mathbb{R}^{J\times C},
\end{equation}
where $J$ is the number of patch tokens and $C$ is the feature dimension. Our goal is to generate pseudo-anomaly features that are not only corrupted at the feature level but also coherent across views.

To achieve this, our framework processes the encoded features in two sequential stages: independent feature perturbation followed by late feature aggregation. First, we apply a stochastic perturbation operator $\mathcal{P}(\cdot)$ independently to each view to inject anomaly-like corruptions:
\begin{equation}
\tilde{F}_i^{(v)}
=
\mathcal{P}\!\left(F_i^{(v)}\right).
\end{equation}
Unfortunately, applying it independently produces scattered, view-inconsistent pseudo-anomaly features. Therefore, we immediately route these perturbed features through a \emph{consensus aggregation} module $\mathcal{A}$ so that complementary multi-view evidence can be jointly incorporated. We concatenate the perturbed features along the channel dimension while preserving their token-grid organization:
\begin{equation}
\bar{F}_i
=
\mathrm{Concat}\!\left(
\tilde{F}_i^{(1)},
\tilde{F}_i^{(2)},
\ldots,
\tilde{F}_i^{(V)}
\right)
\in\mathbb{R}^{J\times VC},
\end{equation}
where features sharing the same token-grid index are concatenated across views. The aggregation module maps the concatenated representation to a $J\times VC$ feature, which is subsequently divided into $V$ view-specific features of dimension $J\times C$:
\begin{equation}
\mathcal{A}\!\left(\bar{F}_i\right)
=
\mathrm{Concat}\!\left(
z_i^{(1)},
z_i^{(2)},
\ldots,
z_i^{(V)}
\right)
\in\mathbb{R}^{J\times VC}.
\end{equation}
The aggregated pseudo-anomaly feature corresponding to the $v$-th view is then defined as
\begin{equation}
z_i^{(v)}
=
\pi_v\!\left(
\mathcal{A}\!\left(\bar{F}_i\right)
\right)
\in\mathbb{R}^{J\times C},
\qquad
v=1,2,\ldots,V,
\end{equation}
where $\pi_v(\cdot)$ denotes the operation that selects the $v$-th channel block from the output of the aggregation module.

By aggregating \emph{after} perturbation, we essentially perform a latent-space consensus operation. While the injected noise is independent and stochastic for each view, the underlying structural features $F_i^{(v)}$ remain geometrically correlated across viewpoints. The aggregation module learns to suppress view-specific stochastic noise while emphasizing the shared structural manifold, thereby producing a pseudo-anomaly that resembles a coherent physical perturbation of the object rather than isolated sensor noise. As illustrated in Figure~\ref{fig:tsne_feature_vis}, features generated independently for each view are widely scattered in the latent space, indicating view-biased anomaly representations. In contrast, after aggregation, the features become highly compact and are drawn closer to their shared centers. This confirms that our consensus aggregation strategy successfully constructs view-aligned anomaly representations that better reflect the coherent defect patterns encountered in real-world multi-view scenarios.

To instantiate the perturbation operator $\mathcal{P}$, we draw inspiration from prior studies on feature-space corruption \cite{you2022unified, liu2023simplenet, guo2025dinomaly, zavrtanik2021draem}. Specifically, we sequentially apply feature jitter, random masking, and dropout. First, feature jitter injects stochastic variations into the latent representation:
\begin{equation}
F_{i,\mathrm{jit}}^{(v)}
=
\mathcal{T}_{\mathrm{jit}}\!\left(F_i^{(v)}\right)
=
F_i^{(v)}
+
\alpha\Delta_i^{(v)},
\end{equation}
where $\Delta_i^{(v)}\in\mathbb{R}^{J\times C}$ denotes a random noise tensor and $\alpha$ controls the jitter magnitude (set to 20). Next, random masking is imposed to further corrupt partial token information:
\begin{equation}
F_{i,\mathrm{mask}}^{(v)}
=
\mathcal{T}_{\mathrm{mask}}\!\left(
F_{i,\mathrm{jit}}^{(v)}
\right).
\end{equation}
Finally, dropout is applied to randomly suppress feature activations:
\begin{equation}
\tilde{F}_i^{(v)}
=
\mathcal{T}_{\mathrm{drop}}\!\left(
F_{i,\mathrm{mask}}^{(v)}
\right).
\end{equation}
By cascading these operations
($\mathcal{P}
=
\mathcal{T}_{\mathrm{drop}}
\circ
\mathcal{T}_{\mathrm{mask}}
\circ
\mathcal{T}_{\mathrm{jit}}$),
we ensure the model encounters highly diverse and robust anomalous patterns during training before the consensus aggregation occurs.

\begin{figure}
\centering
\includegraphics[width=1.0\columnwidth]{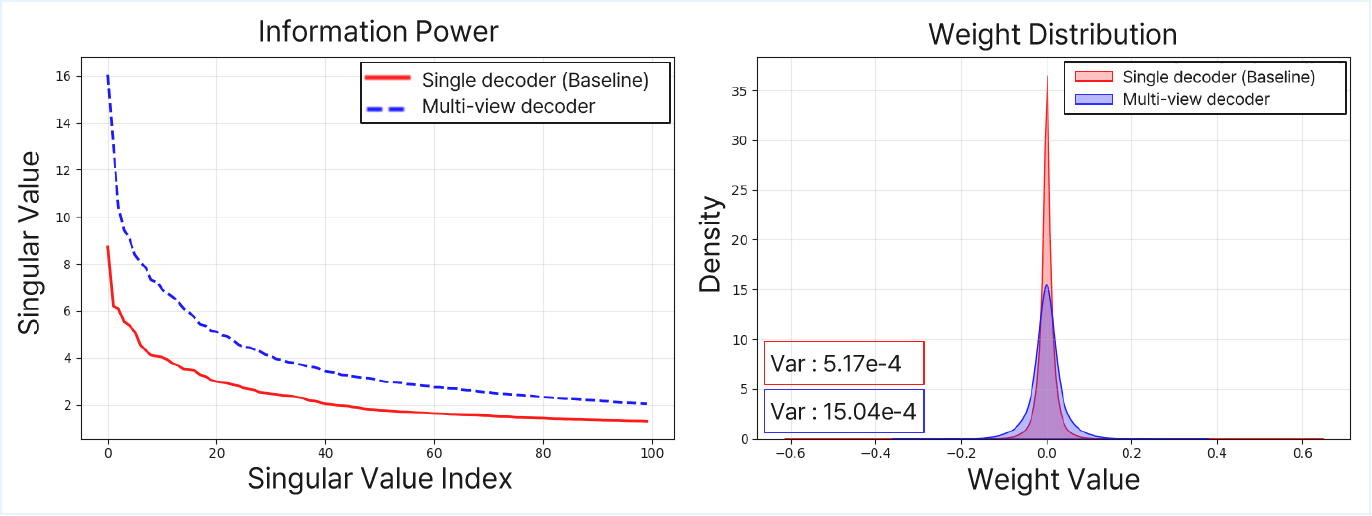}
%\Description={Weight analysis of the baseline decoder and the proposed multi-view decoder.}
\caption{
Comparison of singular values and weight distributions between the baseline single decoder and our proposed multi-view decoder. Higher singular values indicate greater representational capacity, while higher variance in the weight distribution suggests that the decoder learns more expressive and diverse parameters instead of collapsing around zero.
}
\label{fig:weight_analysis}
\end{figure}

\subsection{Fourier-attended view-specialized decoders}
\label{sec:multi_decoder}

To reconstruct view-specific normal features from the aggregated pseudo-anomaly
$z_i^{(v)}$, a single shared decoder often struggles, biasing the model toward
averaged characteristics rather than fine-grained normality patterns. To preserve
these view-dependent details, we propose a view-specialized multi-decoder
architecture. Formally, we define the multi-view decoder as
\begin{equation}
\mathcal{D}
=
\{\mathcal{D}^{(v)}\}_{v=1}^{V},
\end{equation}
where $\mathcal{D}^{(v)}$ denotes the decoder specialized for the $v$-th view.
Given the aggregated pseudo-anomaly feature $z_i^{(v)}$, the $v$-th decoder
produces a reconstructed feature representation
\begin{equation}
\hat{F}_i^{(v)}
=
\mathcal{D}^{(v)}\!\left(z_i^{(v)}\right)
\in\mathbb{R}^{J\times C},
\end{equation}
where $\hat{F}_i^{(v)}$ denotes the reconstructed feature for the $v$-th view,
$H_f$ and $W_f$ denote the height and width of the spatial feature grid,
respectively, $J=H_fW_f$ is the number of spatial tokens, and $C$ is the
feature dimension. The corresponding target feature extracted from the shared
pre-trained encoder is denoted by
\begin{equation}
F_i^{(v)}
=
\mathcal{E}\!\left(I_i^{(v)}\right)
\in\mathbb{R}^{J\times C}.
\end{equation}

While allocating an individual decoder to each view increases representational
capacity, the decoders must still be prevented from trivially copying corrupted
local evidence. If a decoder directly exploits locally available defect patterns
via standard spatial attention, it can easily fall into shortcut behavior that
resembles identity mapping. To mitigate this, we equip each $\mathcal{D}^{(v)}$
with Fourier attention \cite{rao2021gfnet}. By transforming features into the
frequency domain, Fourier attention encourages reconstruction at the level of
global structural patterns rather than direct token-wise copying.

Given an intermediate decoder sequence feature of the $v$-th view,
$H_i^{(v)}\in\mathbb{R}^{J\times C}$, we reshape it into a 2D feature map
\begin{equation}
H_{i,\mathrm{map}}^{(v)}
=
\operatorname{reshape}_{H_f\times W_f\times C}
\!\left(H_i^{(v)}\right)
\in\mathbb{R}^{H_f\times W_f\times C}.
\end{equation}
We then apply a two-dimensional real-input Fourier transform,
\begin{equation}
X_i^{(v)}
=
\mathcal{F}_r\!\left(H_{i,\mathrm{map}}^{(v)}\right)
\in
\mathbb{C}^{H_f\times W_{f,r}\times C},
\end{equation}
where
$W_{f,r}=\lfloor W_f/2\rfloor+1$ and $\mathcal{F}_r$ denotes the
two-dimensional real-input fast Fourier transform. The resulting spectrum is
modulated by a learnable view-specific complex filter as
\begin{equation}
\tilde{X}_i^{(v)}
=
X_i^{(v)}
\odot
\Psi\!\left(K^{(v)};H_f,W_{f,r}\right),
\end{equation}
where
$K^{(v)}\in\mathbb{C}^{H_0\times W_{r,0}\times C}$
denotes the learnable spectral filter specialized for the $v$-th view,
$H_0$ and $W_{r,0}$ denote its initial frequency-grid dimensions,
$\Psi(\cdot;H_f,W_{f,r})$ resizes the filter to the current frequency
resolution, and $\odot$ denotes element-wise multiplication.
The filtered spectrum is subsequently transformed back to the spatial domain:
\begin{equation}
\tilde{H}_{i,\mathrm{map}}^{(v)}
=
\mathcal{F}_r^{-1}\!\left(\tilde{X}_i^{(v)}\right)
\in
\mathbb{R}^{H_f\times W_f\times C}.
\end{equation}
Finally, the spatial feature map is reshaped back into a sequence and passed
through a projection layer and dropout:
\begin{equation}
\hat{H}_i^{(v)}
=
\operatorname{Drop}\!\left(
\operatorname{Proj}\!\left(
\operatorname{reshape}_{J\times C}\!\left(
\tilde{H}_{i,\mathrm{map}}^{(v)}
\right)
\right)
\right)
\in\mathbb{R}^{J\times C}.
\end{equation}

\begin{figure}
\centering
\includegraphics[width=1.0\columnwidth]{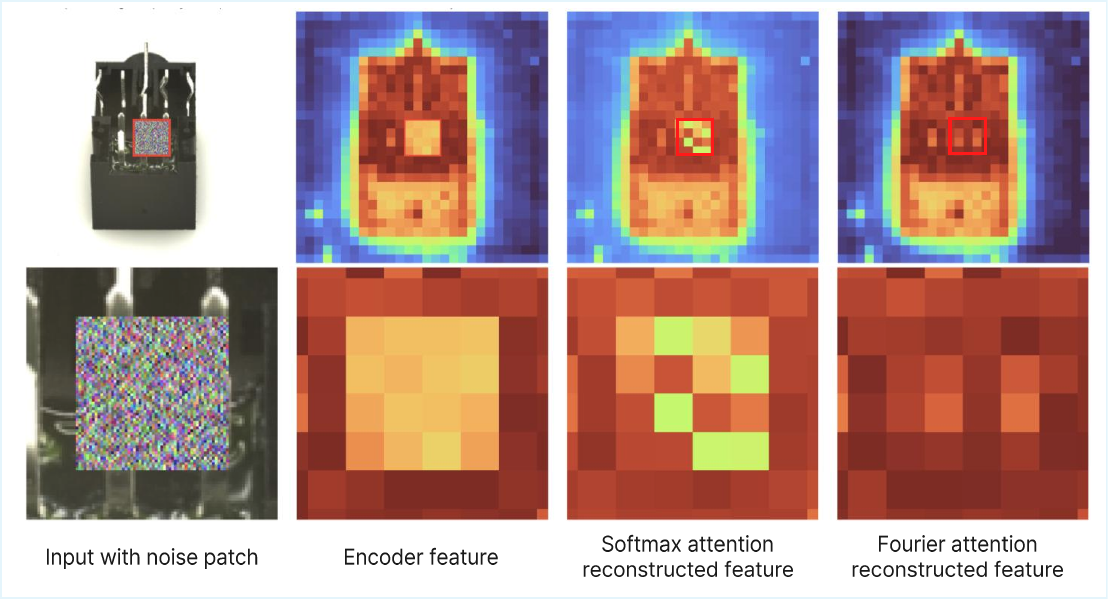} 
%\Description={Feature map visualization}
\caption{
    Visualization of reconstructed feature maps for different attention mechanisms. Successful normal-feature reconstruction should suppress the injected
noise patch.
}
\label{fig:attn_type_visualization}
\end{figure}

The above design acts as a global re-synthesis process. Because local corruptions
affect the global spectrum, the decoder is forced to recover features using
globally consistent structural regularities rather than spatial token copying.
As illustrated in Figure~\ref{fig:attn_type_visualization}, it behaves similarly
to a low-pass filtering process, smoothing out irregular high-frequency anomalies
(e.g., cracks and scratches) and preventing shortcut identity mapping. Furthermore,
as demonstrated in Figure~\ref{fig:weight_analysis}, our view-specialized decoders
exhibit significantly larger singular values and weight variances than a shared
baseline. This confirms that our integrated design possesses the necessary
representational capacity to capture complex view-specific details without
collapsing into averaged characteristics.

\subsection{Training objective}
\label{sec:training}

To optimize the trainable components of PAR, we employ a
global cosine reconstruction loss~\cite{guo2023recontrast}. For the $v$-th view of the $i$-th sample, the reconstruction loss
between the decoder output and the target feature $F_i^{(v)}$ is
defined as
\begin{equation}
\mathcal{L}_{\mathrm{gc},i}^{(v)}
=
1-
\cos\!\,\Bigl(
\mathrm{vec}\!\left(F_i^{(v)}\right),
\mathrm{vec}\!\left(\hat{F}_i^{(v)}\right)
\Bigr),
\end{equation}
where $\mathrm{vec}(\cdot)$ denotes the flattening operation and the target feature $F_i^{(v)}$ is detached from gradient computation. The overall reconstruction loss for the $i$-th sample is obtained by averaging the cosine discrepancy across all $V$ views
\begin{equation}
\mathcal{L}_{\mathrm{rec},i}
=
\frac{1}{V}
\sum_{v=1}^{V}
\mathcal{L}_{\mathrm{gc},i}^{(v)}.
\end{equation}
Minimizing $\mathcal{L}_{\mathrm{rec},i}$ jointly trains the consensus aggregation module and the view-specialized decoders to restore multi-view normal representations, while the shared encoder remains frozen.

\section{Experimental results}

We now describe our experimental setup and present both quantitative
and qualitative results. We also conduct ablation studies to evaluate
the individual contributions of the proposed components.

\begin{table*}[t]
\centering
\caption{Comparison of quantitative experimental results on the Real-IAD dataset. Best value in each row for each metric is boldfaced, and second best is underlined. Scores are reported as S-AUROC / I-AUROC / P-AUROC. }
\label{tab:main_results}
\setlength{\tabcolsep}{3pt}
\renewcommand{\arraystretch}{1.08}
\resizebox{\textwidth}{!}{%
\begin{tabular}{l|ccccc|ccccccc}
\hline
Class & \multicolumn{5}{c|}{Representation-based methods} & \multicolumn{7}{c}{Reconstruction-based methods} \\
\hline
& PatchCore & CFlow & FiCo & VSAD & Multi-Flow & DRAEM & DeSTSeg & RD++ & CKAAD & RealNet & MVAD & PAR~(ours) \\
\hline
Audiojack & \second{93.7}/81.4/97.6 & 86.5/81.1/89.9 & 92.3/89.5/98.3 & 90.3/89.7/98.7 & \best{96.6}/\second{90.7}/97.7 & 91.4/83.2/92.8 & 88.5/81.9/95.6 & 86.4/85.4/86.3 & 87.8/89.2/95.3 & 91.3/79.8/91.4 & 93.1/86.9/\second{98.8} & 93.3/\best{91.9}/\best{99.5} \\
Bottle cap & 94.1/91.7/94.2 & \best{98.9}/86.8/95.3 & \second{98.8}/\best{97.7}/98.6 & 98.2/\second{97.5}/98.6 & 98.0/94.6/99.1 & 96.3/67.6/88.1 & 96.6/85.3/95.1 & 96.4/95.0/98.5 & 95.1/92.8/97.6 & 97.3/92.8/98.5 & \best{98.9}/95.6/\second{99.7} & 97.8/96.9/\best{99.8} \\
Button battery & 81.3/79.2/96.7 & 93.3/77.8/92.0 & 81.9/77.4/91.2 & \second{96.4}/\best{93.8}/97.6 & \best{96.6}/89.4/98.6 & 94.6/84.4/93.1 & 92.0/90.8/97.0 & 94.4/89.0/95.9 & 91.1/85.7/96.8 & 90.6/82.5/96.6 & 93.4/90.8/\second{98.8} & 94.7/\second{93.1}/\best{99.3} \\
End cap & 86.7/80.6/96.2 & 83.2/75.2/89.8 & 89.0/82.7/95.9 & \second{94.7}/89.3/96.8 & 90.9/84.8/95.5 & 75.4/64.8/86.3 & 87.2/81.5/94.0 & 93.7/86.7/96.0 & \best{94.8}/\second{90.0}/96.4 & 81.4/72.8/92.0 & 91.9/85.8/\second{98.1} & 94.4/\best{90.5}/\best{99.4} \\
Eraser & 90.6/90.7/98.3 & 92.1/90.3/98.9 & 89.4/90.1/97.2 & 93.8/\second{95.0}/99.1 & 92.8/92.1/98.8 & 69.4/70.4/81.3 & \second{95.6}/\best{96.3}/98.7 & 93.6/90.4/96.2 & 92.3/91.9/96.3 & 88.9/87.2/98.2 & 91.8/91.2/\second{99.2} & \best{96.3}/94.4/\best{99.7} \\
Fire hood & 87.2/81.6/94.3 & 88.2/83.6/97.4 & 89.2/85.6/96.1 & 90.4/88.3/98.4 & \best{95.5}/\best{93.4}/98.4 & 83.9/72.0/83.6 & 93.4/88.1/95.6 & 89.5/83.8/96.7 & 84.7/82.0/97.9 & 85.6/77.9/98.1 & 90.2/84.6/\second{99.1} & \second{94.7}/\second{90.3}/\best{99.6} \\
Mint & 75.8/70.8/95.3 & 74.5/69.6/94.2 & 65.5/65.9/92.6 & 85.7/\best{85.8}/\second{97.0} & \second{87.8}/79.4/95.3 & 78.6/70.2/83.1 & 80.0/70.6/93.4 & 84.7/82.2/95.9 & 79.4/73.3/95.4 & 73.1/66.6/93.6 & 85.9/79.5/\best{98.5} & \best{89.4}/\second{83.1}/\best{98.5} \\
Mounts & 98.9/85.4/96.4 & 98.3/85.5/98.1 & 96.4/83.1/96.2 & 96.9/\second{88.5}/98.9 & 99.5/84.3/95.8 & 89.3/73.3/84.0 & 96.1/83.4/96.9 & 95.7/\best{88.9}/97.1 & 97.0/85.9/98.6 & 97.2/85.0/\second{99.0} & \second{99.6}/87.9/\second{99.0} & \best{99.9}/87.8/\best{99.5} \\
Pcb & 94.5/94.2/97.3 & 77.4/75.8/94.3 & 89.7/89.5/96.2 & \second{94.6}/\second{94.5}/98.2 & 93.9/91.8/98.7 & 90.5/87.6/95.5 & 91.0/87.7/97.5 & 91.8/91.9/97.7 & 93.9/93.1/97.3 & 86.4/77.0/94.5 & 91.1/91.3/\second{99.4} & \best{96.1}/\best{95.9}/\best{99.7} \\
Phone battery & 93.1/89.6/98.1 & 91.3/86.3/97.1 & 92.7/91.0/98.4 & 94.3/91.1/97.5 & \best{98.7}/90.8/86.2 & \second{97.8}/76.6/85.4 & 93.2/83.6/87.3 & 96.4/91.3/98.3 & 93.9/91.2/98.2 & 86.5/81.4/97.3 & 94.2/\second{92.5}/\second{99.2} & 97.1/\best{94.3}/\best{99.7} \\
Plastic nut & 95.9/88.7/96.4 & 86.7/77.1/95.3 & 94.0/88.9/95.9 & 97.8/\best{95.2}/98.6 & \best{99.0}/88.0/98.7 & 89.8/71.2/80.8 & 94.2/88.1/96.1 & 97.0/92.6/98.5 & 92.1/89.1/98.5 & 87.1/80.3/95.6 & 97.2/91.3/\second{99.6} & \second{98.4}/\second{94.7}/\best{99.8} \\
Plastic plug & 87.7/85.2/96.6 & 90.6/84.9/94.6 & 93.7/89.2/96.1 & \second{95.5}/\best{92.1}/\best{99.3} & 93.9/88.5/97.9 & 85.3/71.8/76.9 & 93.8/83.9/95.6 & 95.3/\second{92.0}/98.6 & 91.5/87.2/98.5 & 86.0/80.3/94.2 & 94.4/89.7/99.0 & \best{97.6}/91.3/\second{99.1} \\
Porcelain doll & 90.8/79.9/96.1 & 95.2/83.4/96.3 & 93.9/83.5/98.3 & 94.6/\second{89.3}/96.2 & 94.4/86.6/98.3 & 94.0/75.7/86.2 & 94.4/81.6/95.8 & \second{97.2}/\best{90.4}/98.2 & 96.8/86.8/97.1 & 90.2/80.1/96.9 & 96.2/87.8/\best{99.1} & \best{98.5}/85.5/\second{98.8} \\
Regulator & 75.7/71.3/96.6 & 66.4/59.6/90.9 & 93.8/88.6/97.9 & 91.5/89.5/98.1 & \second{94.2}/87.7/98.0 & 85.6/72.1/86.0 & 92.2/87.9/97.5 & 91.1/\second{90.1}/98.2 & 83.6/82.4/98.1 & 74.5/65.2/95.5 & 87.7/85.2/\second{99.1} & \best{99.4}/\best{92.0}/\best{99.6} \\
Strip base* & \best{99.8}/\best{99.4}/98.8 & 99.0/97.5/98.2 & \second{99.7}/\best{99.4}/\second{99.6} & \second{99.7}/98.9/98.6 & 99.6/98.2/98.4 & 80.0/87.8/94.9 & 98.9/98.4/98.8 & 99.4/\second{99.3}/\second{99.6} & 99.5/97.9/99.0 & 99.0/96.8/98.1 & 99.6/\best{99.4}/\best{99.7} & 99.5/\best{99.4}/\best{99.7} \\
Sim card set & 93.3/94.3/97.9 & 96.1/95.1/98.5 & 96.2/95.6/\second{98.7} & 99.0/94.2/97.8 & 98.8/90.7/\second{98.7} & \second{99.7}/94.1/96.0 & 97.4/91.7/96.9 & 96.9/95.9/97.6 & 98.7/\second{96.8}/98.0 & 91.2/91.4/97.3 & 98.2/96.2/98.5 & \best{99.8}/\best{97.1}/\best{99.0} \\
Switch & 94.3/93.1/96.9 & 96.0/96.0/97.9 & 96.2/94.8/96.5 & 97.1/\second{96.9}/\second{99.3} & \second{99.3}/96.0/90.3 & 92.6/84.7/89.7 & 96.3/95.0/99.1 & 97.3/96.7/99.2 & 98.6/95.1/97.1 & 87.7/82.7/96.1 & 96.3/93.1/\best{99.5} & \best{99.5}/\best{98.5}/\second{99.3} \\
Tape & 99.2/97.0/98.7 & 99.5/96.6/99.2 & 99.4/96.8/99.3 & 97.4/96.8/99.6 & 97.1/94.3/97.2 & 99.1/91.5/97.4 & 98.6/94.6/99.2 & \second{99.8}/97.9/99.6 & 93.3/93.0/98.7 & 96.8/91.9/99.5 & \best{100.0}/\second{98.1}/\second{99.7} & \second{99.8}/\best{98.3}/\best{99.8} \\
Terminalblock & 96.6/90.5/99.6 & 95.7/88.8/97.5 & 97.3/93.5/99.2 & 98.5/96.9/99.7 & \best{99.1}/96.4/99.3 & 68.9/55.9/83.1 & 96.8/93.1/97.5 & 93.7/96.9/98.4 & 98.4/96.3/99.4 & 92.3/82.9/91.3 & \second{98.6}/\second{97.3}/\second{99.8} & \second{98.6}/\best{98.6}/\best{99.9} \\
Toothbrush & 91.6/85.3/96.8 & 92.6/80.0/92.8 & 90.2/83.2/95.5 & 95.3/86.7/97.0 & \second{96.9}/85.2/95.2 & 93.0/74.6/68.0 & 92.0/\second{88.3}/95.4 & 95.6/85.3/96.9 & 95.7/\best{88.7}/\second{97.2} & 86.8/69.7/92.0 & 95.8/85.5/\best{97.3} & \best{98.9}/86.8/96.0 \\
Toy & 91.5/82.9/95.9 & 70.1/63.2/86.4 & 84.6/78.5/90.4 & 92.6/\best{88.9}/\second{96.1} & \second{96.5}/88.3/87.3 & 68.0/62.0/60.0 & 91.6/82.1/89.7 & 91.5/86.6/95.7 & 91.8/\second{88.8}/95.3 & 70.8/64.0/90.2 & 93.6/86.5/\best{97.3} & \best{98.5}/87.8/94.9 \\
Toy Brick & 79.9/69.8/92.4 & 82.1/81.2/96.4 & 88.5/78.4/96.1 & \second{89.6}/80.4/97.5 & \best{91.7}/\best{87.4}/94.6 & 67.6/65.7/91.2 & 84.8/79.0/95.3 & 82.6/78.4/95.8 & 79.5/77.1/92.3 & 82.4/78.1/94.2 & 80.8/77.9/\second{97.6} & 85.5/\second{82.6}/\best{98.1} \\
Transistor1 & 99.0/94.8/98.3 & 98.1/92.6/96.9 & 97.6/96.3/98.4 & 97.4/95.2/99.1 & 98.6/94.9/93.1 & 93.8/83.1/88.0 & 97.2/95.0/96.7 & 97.8/96.9/97.9 & 98.7/93.8/98.1 & 82.4/78.1/92.3 & \second{99.8}/\second{97.9}/\second{99.5} & \best{99.9}/\best{98.5}/\best{99.6} \\
U Block & 96.8/90.1/96.6 & 94.8/87.0/95.6 & 95.7/89.5/97.4 & 98.4/89.5/\second{99.4} & 96.6/90.5/98.8 & 89.7/73.9/88.6 & 98.1/89.3/98.8 & \second{99.1}/92.4/\second{99.4} & 98.8/92.3/97.5 & 90.8/86.4/96.5 & 98.4/\second{93.1}/\best{99.6} & \best{99.6}/\best{94.5}/\best{99.6} \\
Usb & 88.7/82.4/96.4 & 84.3/80.5/96.0 & 94.5/90.7/97.2 & 93.3/91.3/99.0 & \second{96.8}/\second{93.4}/97.5 & 82.2/72.6/95.7 & 92.5/87.4/97.8 & 94.1/90.3/96.4 & 95.0/92.7/96.9 & 90.0/83.3/97.7 & 96.3/92.8/\second{99.6} & \best{97.1}/\best{95.1}/\best{99.7} \\
USB adaptor & 87.0/79.9/96.8 & 86.7/80.0/94.1 & 86.0/78.6/92.0 & 93.6/85.7/98.1 & 93.4/\best{88.1}/\second{98.7} & \best{94.6}/72.6/85.1 & 93.7/73.9/91.5 & 92.1/82.2/96.6 & 92.2/84.8/97.2 & 84.6/72.0/95.5 & 93.3/83.8/97.3 & \second{94.4}/\second{86.0}/\best{99.0} \\
Vcpill & 88.7/83.5/97.3 & 91.2/89.5/97.7 & 89.8/85.2/95.7 & \second{97.3}/\second{92.8}/98.8 & 95.9/89.5/97.7 & 82.8/75.5/75.9 & 96.6/90.6/98.1 & 96.6/91.1/97.1 & 96.8/90.7/96.7 & 92.4/88.9/98.2 & 95.9/90.8/\second{99.0} & \best{99.3}/\best{95.2}/\best{99.3} \\
Wooden beads & 89.8/86.0/95.8 & 89.4/86.1/96.6 & 90.1/85.7/94.1 & 89.9/87.2/97.2 & \best{94.7}/87.3/97.4 & 86.8/77.8/86.1 & 92.6/86.2/95.8 & 93.0/89.3/98.1 & 90.7/87.7/98.1 & 89.2/82.6/97.9 & 92.7/\second{89.5}/\second{98.6} & \second{94.4}/\best{89.9}/\best{99.2} \\
Woodstick & 87.6/86.3/89.0 & 71.0/79.5/92.8 & 83.8/85.0/96.7 & \second{90.6}/89.1/97.7 & 89.2/88.0/95.5 & 76.4/72.5/90.5 & 87.9/87.9/97.1 & 84.8/85.8/96.4 & 74.3/74.5/92.6 & \best{92.4}/\best{90.6}/95.3 & 85.0/85.7/\second{98.5} & 89.7/\second{89.6}/\best{99.4} \\
Zipper & 98.8/97.9/98.1 & 98.1/95.7/96.6 & 99.8/97.5/98.0 & \second{99.9}/98.9/99.1 & \second{99.9}/98.2/97.5 & 97.8/90.8/82.5 & 99.4/98.5/96.2 & 84.8/85.8/98.3 & 99.8/98.8/98.8 & 99.4/94.0/97.2 & 99.6/\best{99.4}/\second{99.2} & \best{100.0}/\second{99.1}/\best{99.3} \\
\hline
Average & 90.9/86.1/96.5 & 88.9/83.5/95.3 & 91.6/87.7/96.4 & 94.8/\second{91.7}/98.3 & \second{95.9}/90.3/96.5 & 86.5/75.9/85.9 & 93.4/87.4/96.0 & 93.4/90.0/97.2 & 92.5/89.0/97.1 & 88.1/81.4/95.7 & 94.3/90.2/\second{98.9} & \best{96.7}/\best{92.6}/\best{99.1} \\
\hline
\end{tabular}%
}
\end{table*}

\subsection{Experimental settings}

\noindent\textbf{Dataset.} \quad
We evaluate on the Real-IAD \cite{wang2024realiad} dataset, which is a large-scale industrial anomaly detection dataset with multi-view observations over 30 object categories. Following the official split used in our experiments, the training set contains 36,465 normal images, while the test set contains 114,585 images, including 63,256 normal and 51,329 anomalous images.

\noindent\textbf{Compared methods.} We extensively evaluate our proposed framework against a diverse set of state-of-the-art unsupervised anomaly detection baselines. To ensure a comprehensive comparison, we include representation-based approaches (e.g., PatchCore~\cite{roth2022patchcore}, CFlow~\cite{gudovskiy2022cflow}, and VSAD~\cite{chen2026unsupervised}) as well as reconstruction-based paradigms (e.g., DRAEM~\cite{zavrtanik2021draem}, DeSTSeg~\cite{zhang2023destseg}, CKAAD~\cite{fang2025boosting}, RD++~\cite{tien2023rdpp}, and RealNet~\cite{zhang2024realnet}). Most importantly, we benchmark our approach against recent methods specifically tailored for multi-view anomaly detection, namely Multi-Flow~\cite{kruse2025multiflow} and MVAD~\cite{he2026learning}, which represent the current state-of-the-art in this domain.

\noindent\textbf{Evaluation metrics.} \quad
We evaluate our method using sample-level AUROC (S-AUROC), image-level AUROC (I-AUROC), and pixel-level AUROC (P-AUROC). Specifically, S-AUROC evaluates anomaly detection performance at the multi-view sample level by aggregating view-wise scores, I-AUROC measures the ability to distinguish normal and anomalous images, and P-AUROC assesses the accuracy of pixel-wise anomaly localization.
Note that when an anomaly is visible only from a subset of sample images, then the sample is defined as anomalous while some individual images are labeled as ``normal''.

\noindent\textbf{Implementation details.} \quad
We use ViT-Base/14 pre-trained with DINOv2 as the encoder \cite{oquab2024dinov2}. The perturbation module uses a dropout rate of 0.2, a feature jitter scale of 20.0, a jitter application probability of 1.0, and a random patch masking probability of 0.1. We optimize the network with StableAdamW \cite{wortsman2023stable} with AMSGrad \cite{reddi2018convergence} and decoupled weight decay \cite{loshchilov2019decoupled}, using a learning rate of $2\times10^{-3}$, $\beta=(0.9, 0.999)$, and weight decay of $10^{-4}$. The network is trained for 5,000 iterations on Real-IAD. All experiments are conducted on a server equipped with four NVIDIA RTX A6000 GPUs.

\subsection{Quantitative results}
Table~\ref{tab:main_results} summarizes the quantitative results on Real-IAD. Our method achieves the best overall performance among all compared methods, obtaining 96.7\% S-AUROC, 92.6\% I-AUROC, and 99.1\% P-AUROC on average. In particular, PAR improves the average S-AUROC to 96.7\%, compared
with 95.9\% for Multi-Flow and 94.3\% for MVAD. It also achieves
92.6\% I-AUROC and 99.1\% P-AUROC, outperforming both prior
multi-view methods.
These results indicate that our perturb-aggregate-reconstruct framework more effectively leverages complementary multi-view information while preserving view-specific characteristics. The gains are especially meaningful at the sample and image levels, suggesting that our method produces more reliable anomaly evidence across views rather than relying on a shared reconstruction pathway.

\subsection{Qualitative results}
\begin{figure*}
\centering
\includegraphics[width=0.8\textwidth]{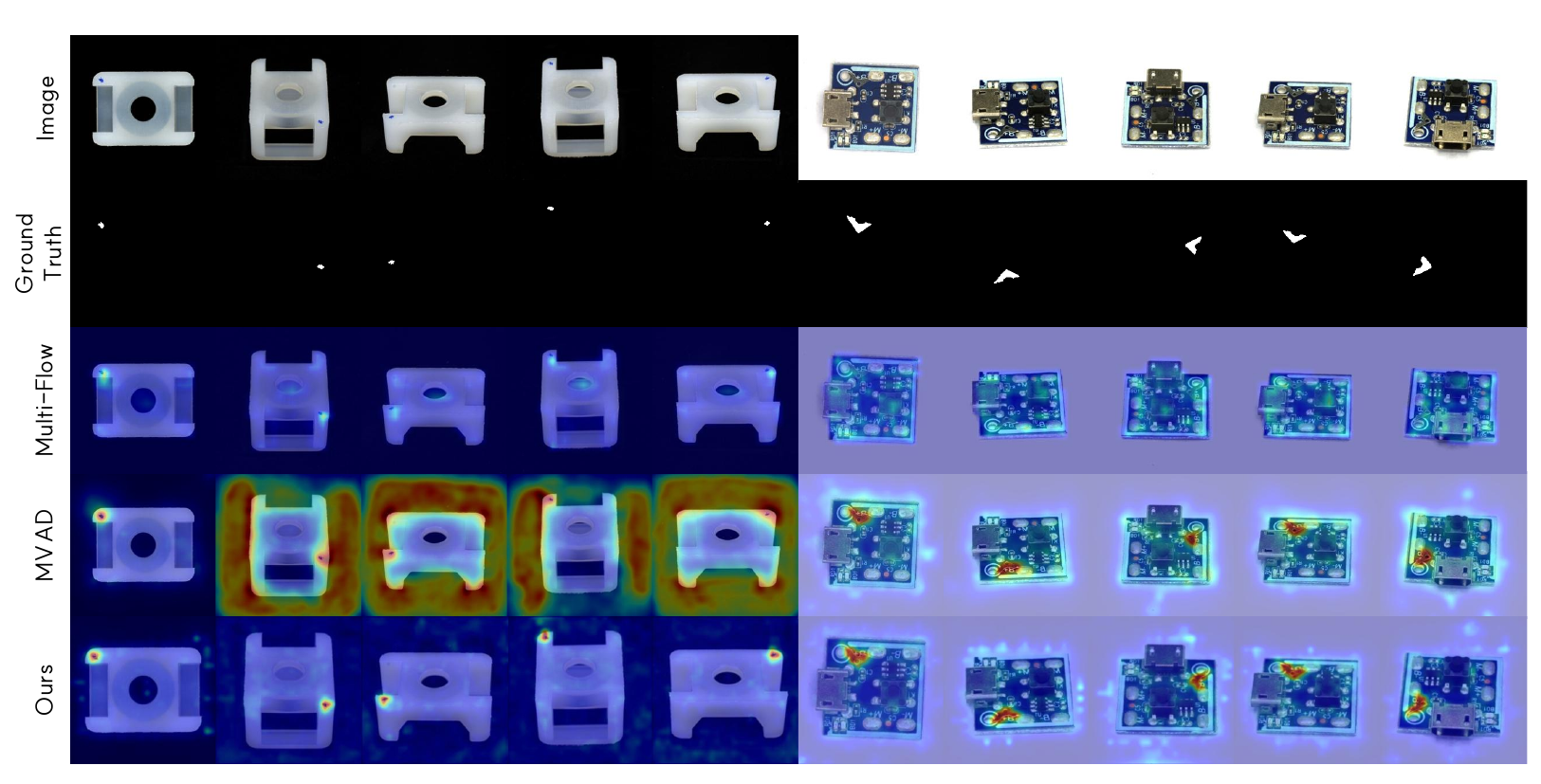} 
%\Description={Visualization of anomaly detection results achieved by SOTA models on the Real-IAD dataset.}
\caption{
     Visualization of anomaly detection results achieved by SOTA models on the Real-IAD dataset.
}
\label{fig:qualitative_results}
\end{figure*}

Figure~\ref{fig:qualitative_results} presents qualitative comparisons with the state-of-the-art multi-view anomaly detection methods, Multi-Flow and MVAD. As shown in the examples, Multi-Flow often produces relatively weak or spatially diffuse responses around the defect regions, while MVAD tends to highlight broader background regions together with the anomalies, resulting in less precise localization. In contrast, our method generates compact and high-confidence anomaly maps that focus more accurately on the defective regions. Moreover, the localized anomaly regions produced by our method more closely resemble the ground-truth masks in both position and shape, demonstrating that PAR is better able to capture the actual extent of defects while suppressing irrelevant activations on normal regions.

\subsection{Ablation study}
We conduct detailed ablation analyses on Real-IAD. In addition,
Table~\ref{tab:component_ablation} reports the component ablation
results on three benchmark datasets to evaluate the generality of
the proposed components. Additional ablation results and experimental results on other datasets are provided in the supplementary material.

\noindent\textbf{Component ablation.}
We conduct component ablations to evaluate the effectiveness of consensus
aggregation, view-specialized decoding, and Fourier attention, as shown in
Table~\ref{tab:component_ablation}. The baseline follows a standard
reconstruction-based framework, in which a pre-trained encoder extracts normal
feature representations and a single decoder reconstructs them from
perturbed features. Consensus aggregation substantially improves
performance by integrating the independently perturbed multi-view features
into more coherent pseudo-anomaly representations. The view-specialized
decoders enable the model to capture view-dependent characteristics that
are essential for multi-view anomaly detection. Fourier attention further
improves reconstruction by preventing trivial identity mapping and
promoting globally consistent structural recovery. Each component
contributes to the overall performance, and their combination yields the
best results, demonstrating their complementary effects.

\noindent\textbf{Perturbation analysis.}
The impact of different pseudo-anomaly generation strategies is analyzed by applying dropout, feature jitter, and random masking to corrupt feature representations prior to multi-view fusion. Detailed results are provided in Table~\ref{tab:feature_perturb_ablation}.
Using a single perturbation provides limited improvement, whereas combining multiple perturbations consistently leads to better performance. In particular, applying all three perturbations achieves the best results, indicating that diverse feature corruption helps the model learn more discriminative anomaly representations.
All other components of the full model are retained while varying
only the perturbation operations.

\noindent\textbf{Attention type analysis.}
Table~\ref{tab:attention_ablation} compares different attention mechanisms within the decoder. Softmax attention yields the lowest performance, while masked and linear attention substantially improve all three metrics, indicating that restricting direct spatial correspondence helps alleviate shortcut reconstruction. Fourier attention achieves the best results, slightly outperforming linear attention. These results support the effectiveness of frequency-domain filtering in promoting globally consistent reconstruction.

\noindent\textbf{Multi-decoder analysis.}
The effect of the proposed multi-decoder design is evaluated by comparing it with a single-decoder baseline, as reported in Table~\ref{tab:decoder_comparison}. For a fair comparison, we also increase the capacity of the single-decoder baseline by using a larger MLP ratio and a deeper feed-forward network (FFN). Despite this, the proposed multi-decoder model consistently achieves the best performance across all metrics, obtaining 96.73 S-AUROC, 92.62 I-AUROC, and 99.12 P-AUROC. In contrast, simply increasing the capacity of a single shared decoder does not improve performance and even leads to slight degradation. These results suggest that the performance gain does not stem from larger model capacity, but from explicitly decomposing the reconstruction process into view-specialized pathways.

\noindent\textbf{Missing view evaluation.}
Table~\ref{tab:missing_view_ablation} evaluates the proposed method under
missing-view scenarios. We progressively remove input views at inference
time while keeping the model trained on full-view data. Compared with
Multi-Flow and MVAD, the proposed method achieves the best performance
for nearly all metrics and missing-view settings while remaining
competitive in the remaining case. Although performance generally
decreases as views are removed, PAR preserves its performance more
effectively than the competing methods. This robustness can be attributed in part to the view-specialized decoders, which preserve view-dependent normality patterns and reduce reliance on a single shared reconstruction pathway.

\begin{table}[h]
\centering
\caption{Component ablation study on three benchmark datasets.
Metrics are reported as S-AUROC / I-AUROC / P-AUROC (\%).
The baseline uses feature perturbation without consensus aggregation,
a single shared decoder, and linear attention. Consensus aggregation,
view-specialized decoding, and Fourier attention are then added
incrementally.}
\label{tab:component_ablation}
\resizebox{\columnwidth}{!}{%
\begin{tabular}{l|ccc}
\toprule
\textbf{Proposed methods} & \textbf{Real-IAD} & \textbf{MANTA} & \textbf{M$^2$AD} \\
\midrule
Baseline      & 78.92 / 74.08 / 95.15 & 63.65 / 60.95 / 79.77 & 67.53 / 61.63 / 87.64 \\
+ Consensus aggregation          & 96.20 / 91.94 / 99.08 & 96.58 / 89.01 / 96.88 & 91.31 / 85.48 / \underline{95.52} \\
+ View-specialized decoders & \underline{96.63} / \underline{92.44} / \underline{99.10} & \underline{96.69} / \underline{89.52} / \underline{97.14} & \underline{91.44} / \underline{87.68} / 95.32 \\
+ Fourier attention              & \textbf{96.73} / \textbf{92.62} / \textbf{99.12} & \textbf{96.78} / \textbf{89.60} / \textbf{97.15} & \textbf{92.84} / \textbf{89.18} / \textbf{95.90} \\
\bottomrule
\end{tabular}%
}
\end{table}

\begin{table}[t]
\caption{Ablation study of combinations of pseudo-anomaly feature generation methods.} 
\centering
\resizebox{\columnwidth}{!}{
\begin{tabular}{ccc|ccc}
\toprule
Random masking & Feature jitter & Dropout & S-AUROC~(\%) & I-AUROC~(\%) & P-AUROC~(\%) \\
\midrule
& & & 95.71 & 91.98 & 98.97 \\
\checkmark & & & 94.98 & 91.46 & 98.26 \\
& \checkmark & & 95.89 & 92.00 & 98.91 \\
& & \checkmark & 96.54 & 92.36 & \underline{99.11} \\
\checkmark & \checkmark & & 96.01 & 92.37 & 98.89 \\
\checkmark & & \checkmark & 96.59 & \textbf{92.54} & 99.01 \\
& \checkmark & \checkmark & \underline{96.67} & \underline{92.49} & \textbf{99.12} \\
\checkmark & \checkmark & \checkmark & \textbf{96.73} & 92.62 & \textbf{99.12} \\
\bottomrule
\end{tabular}
}
\label{tab:feature_perturb_ablation}   
\end{table}

\begin{table}[!ht]
\caption{Comparison with other attention mechanisms on Real-IAD. ``Attn.'' denotes attention.}
\label{tab:attention_ablation}
\centering
\resizebox{\columnwidth}{!}{
\begin{tabular}{cccc}
\toprule
\textbf{Softmax Attn.} & \textbf{Masked Attn.}  & \textbf{Linear Attn.} & \textbf{Fourier Attn.} \\
\midrule
91.91 / 89.41 / 98.65 &
96.24 / 91.35 / 98.73 &
\underline{96.63} / \underline{92.44} / \underline{99.10} &
\textbf{96.73} / \textbf{92.62} / \textbf{99.12} \\
\bottomrule
\end{tabular}
}
\end{table}

\begin{table}[t]
\centering
\caption{Comparison between the single-decoder baseline and the proposed multi-decoder model. `Dim.' denotes dimension, `FFN' denotes feed-forward network. }
\label{tab:decoder_comparison}
\resizebox{\columnwidth}{!}{%
\begin{tabular}{lccccc}
\hline
Model & Hidden Dim. & FFN Depth & S-AUROC (\%) & I-AUROC (\%) & P-AUROC (\%) \\
\hline
\multirow{2}{*}{Single-decoder} & 3072 & 1 & 96.13 & 91.82 & 99.05 \\
 & 6144 & 2 & 95.57 & 91.82 & 99.02 \\
 \hline
Multi-decoder (ours) & 3072 & 1 & \textbf{96.73} & \textbf{92.62} & \textbf{99.12} \\
\hline
\end{tabular}%
}
\end{table}

\begin{table}[t]
\caption{Performance comparison of multi-view anomaly detection methods
under different numbers of missing views. Scores are reported as
S-AUROC / I-AUROC / P-AUROC.}
    \centering
    \resizebox{\columnwidth}{!}{
        \begin{tabular}{c|ccccc}
            \hline
             \multirow{2}{*}{Method} & \multicolumn{5}{c}{Number of missing views} \\ \cline{2-6} 
            & $0$ & $1$ & $2$ & $3$ & $4$ \\ \hline
 Multi-Flow & 95.85/90.27/96.40 & 57.01/54.00/79.86 & 56.90/52.03/71.48 & 55.67/48.91/62.26 & 54.96/48.57/56.31  \\
MVAD & 94.33/89.56/98.58 & \textbf{92.65}/89.33/98.52 & 91.36/89.06/98.49 & 88.93/88.28/98.29 & 88.92/88.93/98.48 \\
    Ours & \textbf{96.73}/\textbf{92.62}/\textbf{99.12} & 92.45/\textbf{89.96}/\textbf{98.84} & \textbf{91.56}/\textbf{89.97}/\textbf{98.85} & \textbf{90.35}/\textbf{89.56}/\textbf{98.84} & \textbf{91.71}/\textbf{91.71}/\textbf{98.85}  \\
            \hline
        \end{tabular}
        }
    \label{tab:missing_view_ablation}
\end{table}

\section{Conclusion}
In this paper, we presented PAR, a novel reconstruction-based framework tailored for multi-view anomaly detection. Following a perturb-aggregate-reconstruct paradigm, our method first generates highly view-aligned pseudo-anomalies through stochastic feature perturbation and a late consensus aggregation module. These aligned representations are then processed by view-specialized decoders, which effectively increase the model's representational complexity to preserve fine-grained, view-dependent normality patterns. Furthermore, we integrated Fourier attention into the decoders to shift the reconstruction process to the frequency domain, successfully preventing trivial identity mapping and ensuring robust, anomaly-free recovery. By jointly encouraging spatial coherence of pseudo-anomalies, improving view-specific normal-feature reconstruction and preventing identity mapping, PAR achieves improved multi-view anomaly detection performance over state-of-the-art methods on the large-scale Real-IAD benchmark.

\noindent\textbf{Limitations.} Despite its strong performance, PAR has a few limitations. First, the framework assumes a fixed multi-view camera setup with dedicated decoder assignments, which may be less suitable for scenarios with varying or unordered viewpoints. Second, while our consensus aggregation strategy produces highly aligned pseudo-anomalies, it does not strictly enforce geometric consistency of pseudo-anomaly features.
Finally, scaling the number of decoders linearly with the number of views introduces additional memory overhead compared to a single shared decoder.

\begin{acks}
This work was supported by the Korea Evaluation Institute of Industrial Technology (KEIT, RS-2026-25525507, 75\%), Institute of Information \& Communications Technology Planning \& Evaluation (IITP) grant funded by the Korea government(MSIT) (No. RS-2020-II201373, Artificial Intelligence Graduate School Program (Hanyang University), 25\%) and Hanyang University (HY-202500000003991). 
\end{acks}

\clearpage
\bibliographystyle{ACM-Reference-Format}
\bibliography{sample-base}

% =========================================================
% Supplementary Material
% =========================================================

\clearpage

\setcounter{section}{0}
\setcounter{subsection}{0}
\setcounter{table}{0}
\setcounter{figure}{0}

\renewcommand{\thetable}{A\arabic{table}}
\renewcommand{\thefigure}{A\arabic{figure}}

\twocolumn[
\begin{minipage}{\textwidth}
\centering

{\Huge\bfseries
Perturb, Aggregate then Reconstruct: Consensus Aggregation and
Fourier-Attended Multi-Decoders for Reconstruction-based
Multi-View Anomaly Detection\par}

\vspace{0.8em}

{\LARGE\bfseries Supplementary Document\par}

\vspace{1.8em}
\end{minipage}
]

\section{Additional ablation studies}

All additional ablation studies in this section are conducted on the Real-IAD dataset.

\noindent\textbf{Pre-trained backbone analysis.} \quad
The quality of the frozen foundation encoder has a direct impact on unsupervised multi-view anomaly detection performance. To examine this effect, we compare several representative self-supervised backbones that are trained with different pre-training objectives, including masked image modeling (MIM), contrastive learning (CL), and their combination. D-iGPT~\cite{ren2023rejuvenating} is a pure MIM-based model, where the pretext task focuses on recovering target features from masked inputs. MoCo v3~\cite{chen2021empirical} represents a contrastive learning paradigm that learns discriminative representations by pulling semantically similar samples closer and pushing dissimilar ones apart. In contrast, iBOT~\cite{zhou2021ibot} and DINOv2~\cite{oquab2024dinov2} combine contrastive learning and masked image modeling, providing both discriminative and reconstruction-aware supervisory signals during pre-training. As shown in Table~\ref{tab:pretrained_backbones}, encoders trained with both CL and MIM generally provide stronger performance than those relying on a single objective, with DINOv2 achieving the best overall results. This suggests that jointly learning semantic discrimination and masked reconstruction yields feature representations that are more suitable for our reconstruction-based multi-view anomaly detection framework.

\begin{table}[h]
\centering
\caption{Comparison across different pre-trained backbones. "MIM" denotes masked image modeling, and "CL" denotes contrastive learning.}
\label{tab:pretrained_backbones}
\resizebox{\columnwidth}{!}{%
\begin{tabular}{lcccc}
\hline
Pre-trained backbone & Type & S-AUROC (\%) & I-AUROC (\%) & P-AUROC (\%) \\
\hline
D-iGPT~\cite{ren2023rejuvenating}  & MIM    & \underline{96.58} & 92.16 & 98.54 \\
MoCo v3~\cite{chen2021empirical} & CL     & 96.02 & 91.56 & 98.50 \\
iBOT~\cite{zhou2021ibot}    & CL + MIM & 96.49 & \textbf{92.64} & \underline{98.83} \\
DINOv2~\cite{oquab2024dinov2}  & CL + MIM & \textbf{96.73} & \underline{92.62} & \textbf{99.12} \\
\hline
\end{tabular}%
}
\end{table}

\noindent\textbf{Aggregation strategy analysis.} \quad
The effect of different aggregation strategies is summarized in Table~\ref{tab:aggregation_ablation}. Among the compared methods, the simple MLP layer consistently achieves the best performance across all three metrics, outperforming both MLP-Mixer~\cite{tolstikhin2021mlp} and cross-attention~\cite{vaswani2017attention}. While MLP-Mixer shows competitive S-AUROC and cross-attention achieves relatively strong I-AUROC and P-AUROC, neither surpasses the MLP layer overall. These results suggest that, for the proposed perturb-then-aggregate design, a lightweight learned fusion module is sufficient to capture complementary cross-view cues while preserving the multi-view features. 
% In contrast, more complex aggregation strategies do not necessarily provide additional benefit in this setting.

\begin{table}[ht]
\centering
\caption{Ablation study on aggregation strategies.}
\label{tab:aggregation_ablation}
\resizebox{\columnwidth}{!}{%
\begin{tabular}{lccc}
\hline
Aggregation strategy & S-AUROC (\%)& I-AUROC (\%)& P-AUROC (\%) \\
\hline
MLP-Mixer~\cite{tolstikhin2021mlp} & \underline{96.45} & 92.08 & 98.94 \\
Cross-attention~\cite{vaswani2017attention} & 96.18 & \underline{92.24} & \underline{98.96} \\
MLP~\cite{rumelhart1986learning} & \textbf{96.73} & \textbf{92.62} & \textbf{99.12} \\
\hline
\end{tabular}%
}
\end{table}

\noindent\textbf{Consensus aggregation analysis.} \quad
To evaluate whether consensus aggregation improves cross-view consistency
of pseudo-anomaly features, we measure mutual information and cross-view
token retrieval accuracy on Real-IAD.
We use a lightweight two-layer MLP for view mixing, implemented as
Linear$(5 \rightarrow 10)$, GELU, and Linear$(10 \rightarrow 5)$ in the
5-view setting.
For the mutual information analysis, we use the center-region tokens visible
across all views.
Specifically, we estimate mutual information under a bivariate Gaussian
assumption based on cross-view feature correlations.
For a pair $(p,q)$, the cross-covariance is
$\Sigma_{pq}=\frac{1}{M-1}(Z^{(p)}-\mu^{(p)})^\top
(Z^{(q)}-\mu^{(q)})$, where $\mu^{(p)}$, $\mu^{(q)}$, and $M$ denote
the mean feature vectors and the number of shared tokens.
Table~\ref{tab:mutual_info} shows that independent perturbation without
consensus aggregation yields consistently low mutual information.
We further measure cross-view token retrieval accuracy on the same shared
center-region tokens. Table~\ref{tab:token_retrieval} shows that consensus
aggregation consistently improves Top-1 retrieval accuracy across all
view pairs.

\begin{table}[ht]
\centering
\caption{Mutual information scores based on cross-covariance. ``CA'' denotes consensus aggregation.}
\label{tab:mutual_info}
\resizebox{\columnwidth}{!}{%
\begin{tabular}{lccccc}
\hline
&
\shortstack[l]{View 1--View 2} &
\shortstack[l]{View 1--View 3} &
\shortstack[l]{View 1--View 4} &
\shortstack[l]{View 1--View 5} &
Mean \\
\hline
w/o CA
& 0.1015
& 0.0419
& 0.0997
& 0.1122
& 0.0888 \\
\hline
w/ CA
& \textbf{0.2328}
& \textbf{0.1226}
& \textbf{0.2575}
& \textbf{0.2167}
& \textbf{0.2074} \\
\hline
\end{tabular}%
}
\end{table}

\begin{table}[ht]
\centering
\caption{Cross-view token retrieval accuracy (\%) using View 1
as the anchor. ``CA'' denotes consensus aggregation.}
\label{tab:token_retrieval}
\resizebox{\columnwidth}{!}{%
\begin{tabular}{lcccc}
\hline
&
\shortstack[l]{View 1--View 2} &
\shortstack[l]{View 1--View 3} &
\shortstack[l]{View 1--View 4} &
\shortstack[l]{View 1--View 5} \\
\hline
w/o CA
& 9.22
& 2.32
& 7.36
& 2.94 \\
\hline
w/ CA
& \textbf{31.22}
& \textbf{11.99}
& \textbf{37.35}
& \textbf{15.79} \\
\hline
\end{tabular}%
}
\end{table}

\noindent\textbf{Missing-view analysis.} \quad
To examine how PAR performs when some input views are unavailable, we
evaluate the model under missing-view settings.
Missing views are simulated only at inference time by zeroing the
corresponding view features. As shown in Table~\ref{tab:missing_view},
PAR exhibits only modest performance degradation under any single-view
removal, maintaining 95.0--96.5\% S-AUROC, 92.1--92.6\% I-AUROC, and
98.9--99.0\% P-AUROC.

\begin{table}[t]
\centering
\caption{Performance under missing-view settings.
Scores are reported as S-AUROC / I-AUROC / P-AUROC (\%).}
\label{tab:missing_view}
\resizebox{\columnwidth}{!}{%
\begin{tabular}{l|ccccc}
\hline
&
\shortstack[c]{Missing View 1} &
\shortstack[c]{Missing View 2} &
\shortstack[c]{Missing View 3} &
\shortstack[c]{Missing View 4} &
\shortstack[c]{Missing View 5} \\
\hline
Ours
& 96.5 / 92.6 / 99.0
& 95.7 / 92.3 / 99.0
& 95.6 / 92.2 / 98.9
& 95.0 / 92.1 / 98.9
& 95.6 / 92.1 / 98.9 \\
\hline
\end{tabular}%
}
\end{table}

\noindent\textbf{Robustness to affine perturbations.} \quad
To evaluate robustness to moderate spatial misalignment, we apply random
affine perturbations with rotations of $\pm 3^\circ$ and translations of
$\pm 5\%$, as shown in Table~\ref{tab:affine_transformation}. Although the
performance decreases under affine perturbations, PAR maintains strong
results and exhibits substantially greater robustness than MVAD.

\begin{table}[t]
\centering
\caption{Comparison under affine transformations.
Scores are reported as S-AUROC / I-AUROC / P-AUROC (\%).}
\label{tab:affine_transformation}
\begin{tabular}{l|cc}
\hline
&
\shortstack[c]{No transformation} &
\shortstack[c]{Affine transformation} \\
\hline
MVAD
& 90.08 / 86.42 / 97.71
& 63.24 / 63.67 / 90.41 \\
\hline
Ours
& \textbf{96.73 / 92.62 / 99.12}
& \textbf{89.59 / 89.73 / 98.03} \\
\hline
\end{tabular}
\end{table}

\noindent\textbf{Model efficiency analysis.} \quad
Table~\ref{tab:efficiency_comparison} compares the computational
cost, GPU memory usage, and inference throughput of PAR with
Multi-Flow and MVAD under identical inference settings. Although
PAR incurs higher computational cost and memory usage due to its
view-specialized decoders, it achieves a throughput of 102.41 FPS.
All measurements were conducted on an AMD Ryzen 9 7900X CPU
and an NVIDIA RTX 4090 GPU using an input resolution of
$392\times392$ and a batch size of 1. FPS is measured per
multi-view sample, including all input views, and GPU memory
denotes the peak allocated memory during inference.

\begin{table}[t]
\centering
\caption{Model efficiency analysis.}
\label{tab:efficiency_comparison}
\begin{tabular}{lccc}
\hline
 & GFLOPs & GPU Memory (MB) & FPS \\
\hline
Multi-Flow & 36.896 & 315.97 & 369.58 \\
\hline
MVAD & 113.828 & 340.41 & 283.14 \\
\hline
Ours & 1085.83 & 3554.24 & 102.41 \\
\hline
\end{tabular}
\end{table}

\section{Additional datasets}
To further examine the generalization ability of the proposed method, we conduct additional experiments on two large-scale multi-view anomaly detection datasets, namely M$^2$AD and MANTA. 

M$^2$AD is a benchmark designed to evaluate visual anomaly detection under the complex interplay of viewpoint and illumination. It contains 119,880 high-resolution images collected from 999 specimens across 10 categories, where each specimen is captured under 12 synchronized views and 10 illumination settings, resulting in 120 imaging configurations per object. The dataset includes 69,070 normal images and 50,810 anomalous images with pixel-precise annotations, making it a challenging benchmark for multi-view anomaly detection under realistic imaging variations \cite{cao2026visual}. 

MANTA is a large-scale multi-view anomaly detection dataset for tiny industrial objects. Its visual component comprises more than 137.3K images across 38 product categories spanning five domains, among which about 8.6K anomalous images are provided with pixel-level annotations. Each object is captured from five distinct viewpoints, enabling comprehensive multi-view evaluation in fine-grained industrial inspection scenarios \cite{fan2025manta}. 

For both datasets, we use the same training hyperparameters as those
used on Real-IAD, with the number of view-specialized decoders adjusted
to the number of input views. The quantitative comparisons are reported
in Tables~\ref{tab:m2ad_results} and~\ref{tab:manta_main_results}. The baseline results for M$^2$AD and MANTA are
taken from \cite{yan2026pico} and \cite{chen2026unsupervised},
respectively. PAR achieves strong performance on both datasets,
demonstrating that the proposed perturb-aggregate-reconstruct framework
generalizes beyond Real-IAD.

\begin{table*}[!t]
\centering
\caption{Comparison on the M$^2$AD dataset. The best score for each metric in each row is boldfaced, and the
second-best score is underlined. Scores are reported as S-AUROC / I-AUROC / P-AUROC. P-AUROC has not been reported for the CDO model~\cite{cao2023collaborative}.}
\label{tab:m2ad_results}
\setlength{\tabcolsep}{5pt}
\renewcommand{\arraystretch}{1.12}
\resizebox{\textwidth}{!}{%
\begin{tabular}{l|cccccccc}
\toprule
Category & CDO~\cite{cao2023collaborative} & RD++~\cite{tien2023revisiting} & MSFlow~\cite{zhou2024msflow} & INP-Former~\cite{luo2025exploring} & RealNet~\cite{zhang2024realnet} & CKAAD~\cite{fang2025boosting} & MVAD~\cite{he2026learning} & Ours \\
\midrule
Bird
& 73.8/71.8/--
& \best{90.8}/71.3/70.2
& 86.8/61.8/62.0
& 87.7/71.8/67.2
& \second{90.6}/\second{76.3}/\second{97.1}
& 77.6/\second{76.3}/96.3
& 82.2/71.8/97.0
& 85.6/\best{81.6}/\best{97.6} \\
Car
& 84.1/75.7/--
& \second{86.3}/68.6/68.2
& 71.5/69.7/55.9
& 85.6/\second{80.6}/53.9
& 79.0/65.2/\second{92.3}
& 77.6/75.4/89.8
& 84.5/77.8/91.9
& \best{90.3}/\best{89.0}/\best{93.1} \\
Cube
& \best{95.1}/\best{91.1}/--
& 83.2/76.0/74.6
& 80.1/74.6/57.8
& \second{89.4}/\second{86.5}/74.5
& \second{94.5}/81.9/90.1
& 81.1/74.5/90.9
& 82.2/75.4/\second{91.7}
& 88.7/80.7/\best{94.5} \\
Dice
& 76.0/71.8/--
& \best{98.5}/\second{90.0}/89.4
& 79.1/77.5/69.4
& 72.5/74.4/83.7
& 97.3/88.6/\second{94.0}
& \second{98.4}/86.2/92.6
& 97.7/88.0/92.6
& 96.5/\best{90.7}/\best{96.2} \\
Doll
& \best{99.7}/\best{90.4}/--
& 67.4/66.0/65.9
& 57.3/56.2/55.1
& \second{99.0}/\second{84.4}/73.7
& 79.3/70.0/95.3
& 77.0/67.2/96.4
& 76.2/69.9/\second{97.2}
& 82.2/73.5/\best{97.8} \\
Holder
& 96.5/91.7/--
& \second{99.1}/87.8/87.8
& 97.9/68.6/76.6
& 80.5/70.0/76.4
& 98.3/84.6/91.7
& 97.2/\second{94.4}/92.3
& 98.2/90.2/\second{93.0}
& \best{99.4}/\best{97.2}/\best{96.4} \\
Motor
& 93.8/90.5/--
& 92.2/87.9/87.9
& 77.3/61.7/61.4
& 87.6/84.0/61.1
& 88.2/78.4/98.6
& 92.5/88.5/98.5
& \second{95.1}/\second{91.8}/\second{98.9}
& \best{98.1}/\best{97.0}/\best{99.1} \\
Ring
& 86.8/74.0/--
& \best{96.7}/\best{91.6}/\second{90.9}
& 86.9/80.3/72.4
& 69.8/60.8/71.7
& \second{93.0}/88.8/\best{95.8}
& 89.1/81.7/85.7
& 92.4/86.3/88.3
& 91.9/\second{90.3}/\best{95.8} \\
Teapot
& \best{100.0}/\best{96.4}/--
& 91.3/86.0/86.0
& 72.1/65.4/63.9
& 82.1/80.2/79.1
& 99.3/88.9/\best{98.8}
& 99.1/86.5/96.5
& \second{99.6}/90.2/\second{98.5}
& 98.5/\second{95.5}/98.0 \\
Tube
& \second{95.9}/\second{89.0}/--
& 92.1/81.2/81.2
& 80.7/68.8/67.3
& 87.4/77.0/64.1
& 84.0/78.3/\best{95.0}
& 89.7/82.9/90.4
& 94.9/81.8/\second{90.5}
& \best{97.1}/\best{96.4}/\second{90.5} \\
\midrule
Average
& 90.2/\second{84.2}/--
& 89.7/80.6/80.2
& 79.0/68.5/64.2
& 84.2/77.0/70.5
& \second{90.3}/80.1/\second{94.9}
& 87.9/81.4/93.0
& \second{90.3}/82.3/94.0
& \best{92.8}/\best{89.2}/\best{95.9} \\
\bottomrule
\end{tabular}%
}
\end{table*}

\begin{table*}[!t]
\centering
\caption{Comparison on the MANTA dataset. The best score for each metric in each row is boldfaced, and the second-best score is underlined. Scores are reported as S-AUROC / I-AUROC / P-AUROC. 
The starred names abbreviate Round Button Cap, Square Button Cap, and Yellow Green Washer, respectively.
}
\label{tab:manta_main_results}
\setlength{\tabcolsep}{3pt}
\renewcommand{\arraystretch}{1.08}
\resizebox{\textwidth}{!}{%
\begin{tabular}{l|ccccc|ccccccc}
\hline
Class & \multicolumn{5}{c|}{Embedding-based methods} & \multicolumn{7}{c}{Reconstruction-based methods} \\
\hline
& PatchCore & CFlow & FiCo & VSAD & Multi-Flow & DRAEM & DeSTSeg & RD++ & CKAAD & RealNet & MVAD & Ours \\
\hline
Block Inductor & 92.1/\second{92.5}/97.8 & 84.3/87.2/95.7 & 77.7/69.2/87.5 & \second{93.2}/\best{93.7}/\second{98.6} & 89.9/82.3/95.1 & 85.6/81.8/71.8 & 92.1/91.4/97.7 & 91.0/91.4/96.3 & 89.5/91.2/97.2 & 76.1/77.4/88.6 & 90.3/92.0/98.3 & \best{95.5}/90.4/\best{98.8} \\
Copper Standoff & 97.1/97.5/97.9 & 98.0/94.5/96.8 & 66.2/83.7/97.5 & 96.3/\second{97.9}/98.1 & 97.0/91.7/92.8 & 66.9/61.7/71.7 & 91.8/92.8/94.7 & 98.5/\best{98.0}/98.4 & 98.2/\second{97.9}/98.1 & 92.9/87.1/95.3 & \second{99.2}/96.6/\second{98.6} & \best{99.8}/96.4/\best{99.2} \\
Flat Nut & 95.5/95.6/98.2 & 84.8/88.7/97.3 & 83.0/86.2/95.9 & \second{96.4}/\best{97.7}/\best{99.6} & 86.4/77.9/98.2 & 60.1/62.5/69.1 & 84.7/83.2/96.2 & 95.0/\second{96.4}/98.8 & 89.8/92.6/96.2 & 83.8/81.4/94.5 & 90.5/92.9/99.2 & \best{98.3}/89.1/\second{99.4} \\
Led & 96.9/97.8/98.6 & 97.9/97.8/99.1 & 98.8/98.2/98.6 & \second{99.0}/\best{99.8}/99.1 & 92.3/83.2/96.1 & 92.0/90.1/86.9 & 96.0/92.0/97.5 & 98.9/\second{99.0}/\second{99.4} & 97.9/97.2/98.5 & 90.8/92.6/97.3 & 98.6/98.9/99.3 & \best{99.9}/92.8/\best{99.6} \\
Led Pad & 98.7/97.1/97.7 & 91.4/94.0/97.1 & 98.8/96.7/98.5 & 95.0/95.4/97.2 & 89.9/81.1/94.8 & 60.7/58.8/71.7 & 98.3/97.0/95.2 & 96.4/97.1/96.9 & 98.9/\second{97.2}/98.8 & 77.1/78.3/91.8 & \second{99.3}/\best{99.2}/\second{98.9} & \best{99.6}/89.5/\best{99.2} \\
Long Button & 96.0/96.5/98.0 & 92.1/92.5/95.8 & 95.5/96.1/97.8 & \second{98.3}/\best{97.9}/\second{98.6} & 96.2/88.7/96.6 & 83.8/75.8/79.4 & 95.4/97.3/98.5 & 96.3/96.9/98.1 & 98.2/\second{97.8}/98.5 & 85.5/83.9/93.1 & 97.9/97.7/98.5 & \best{98.6}/93.0/\best{98.8} \\
Power Inductor & 81.5/82.3/94.5 & 81.1/80.7/93.7 & 80.9/83.7/96.6 & 86.7/88.3/95.6 & 81.5/75.0/89.3 & 65.1/66.3/76.9 & 87.9/\second{88.7}/94.7 & 83.8/86.7/97.0 & 85.1/86.9/95.6 & 67.1/62.9/87.4 & \second{88.9}/\best{89.6}/\second{97.2} & \best{98.4}/86.9/\best{98.6} \\
Short Button & 95.4/94.8/98.9 & 87.9/87.4/96.9 & 93.1/92.8/98.6 & \second{96.9}/\best{96.8}/\second{99.4} & 96.4/87.1/97.6 & 75.5/67.0/76.1 & 96.2/95.5/99.2 & 94.7/95.3/97.0 & 95.2/95.1/98.1 & 75.2/74.2/92.8 & 95.6/\second{95.8}/99.1 & \best{97.1}/89.6/\best{99.5} \\
Thin Resistor & 97.2/96.0/97.6 & 95.0/95.6/98.0 & 97.8/90.3/94.3 & \second{98.8}/\best{97.9}/\best{99.4} & 96.0/87.7/95.6 & 90.1/87.8/76.2 & 98.3/94.8/96.9 & 98.1/97.0/98.0 & 97.0/\second{97.5}/98.1 & 89.2/88.6/95.5 & 97.4/97.4/97.9 & \best{99.7}/92.7/\second{98.2} \\
Type C & 97.3/95.6/96.8 & 92.6/92.8/97.7 & 98.5/97.4/\second{98.9} & 95.3/96.8/97.6 & 94.5/88.5/87.0 & 89.3/81.9/82.7 & 97.1/97.2/98.3 & 98.0/\second{97.8}/\second{98.9} & 96.8/97.0/98.1 & 84.0/83.7/93.3 & \second{99.1}/\best{98.0}/\best{99.0} & \best{99.9}/95.3/\best{99.0} \\
Wafer Resistor & 94.0/95.6/96.2 & 92.4/93.0/95.3 & 92.5/93.2/98.7 & 97.6/95.6/\second{99.2} & 95.8/82.6/93.3 & 84.0/83.4/71.6 & 96.8/94.6/96.0 & 96.1/\second{96.0}/99.1 & 95.3/94.7/97.4 & 80.9/83.5/95.1 & \best{99.1}/\best{98.0}/99.0 & \second{98.7}/87.1/\best{99.4} \\
\hline
Maize & \second{85.6}/\second{84.1}/\best{92.5} & 68.6/69.7/82.1 & 61.2/67.0/89.2 & 85.1/82.9/91.7 & 69.6/64.4/85.9 & 80.8/79.8/67.8 & 84.4/81.4/79.6 & 75.6/76.6/90.7 & 83.1/82.4/87.3 & 70.7/71.2/83.2 & 80.7/80.5/90.8 & \best{90.4}/\best{85.6}/\second{92.4} \\
Paddy & \second{89.1}/\second{88.5}/\best{85.7} & 86.3/81.6/73.7 & 75.1/73.3/77.3 & 84.3/88.4/\second{83.2} & 88.4/81.9/81.1 & 67.3/66.8/79.7 & 86.6/80.6/63.1 & 87.3/84.8/78.3 & 86.8/84.3/79.6 & 88.0/80.9/75.4 & 88.8/86.0/78.8 & \best{93.9}/\best{92.2}/\best{85.7} \\
Soybean & 88.7/91.1/92.2 & 81.4/86.2/88.2 & 65.4/77.8/90.0 & 92.7/\best{94.5}/\second{93.3} & 89.4/81.0/89.5 & 78.8/83.0/68.2 & 90.7/89.7/85.4 & 87.5/90.9/92.3 & 91.8/\second{94.4}/93.2 & 84.6/84.6/85.8 & \second{93.0}/93.3/92.3 & \best{96.8}/92.6/\best{94.6} \\
Wheat & 86.1/86.9/94.6 & 85.6/86.2/94.4 & 78.7/84.7/94.1 & 86.9/87.6/\second{96.4} & 86.6/78.6/92.6 & 61.0/67.7/82.0 & 86.0/82.7/89.0 & 86.4/85.8/93.5 & 85.2/84.7/93.0 & 80.1/76.9/91.5 & \second{93.5}/\best{92.4}/95.3 & \best{95.2}/\second{88.6}/\best{96.9} \\
\hline
Capsule & 98.9/98.1/93.3 & 96.5/96.8/84.7 & 98.1/98.2/94.1 & \best{99.4}/\second{98.9}/\best{95.5} & 98.5/95.5/91.1 & 98.8/98.0/84.6 & 98.8/98.1/89.3 & \second{99.0}/98.4/90.2 & 97.4/96.9/87.5 & 95.7/91.9/79.8 & \second{99.0}/\best{99.4}/90.2 & 98.3/96.9/\second{94.3} \\
Coated Tablet & 99.2/98.9/\best{99.8} & 98.1/97.4/\second{99.7} & 98.3/98.8/98.9 & 97.4/97.1/\second{99.7} & 95.9/89.3/99.5 & 98.6/97.6/99.0 & \best{99.7}/\best{99.5}/\best{99.8} & 98.2/98.6/\second{99.7} & \second{99.6}/\second{99.3}/\second{99.7} & 98.7/98.3/99.4 & 99.3/99.2/\best{99.8} & \second{99.6}/96.6/\best{99.8} \\
Embossed Tablet & 95.3/96.1/97.6 & 90.0/88.9/92.3 & 95.6/95.0/98.3 & 97.8/96.1/\second{98.5} & 90.5/83.9/\second{98.5} & 83.2/79.6/77.8 & 95.3/94.7/93.5 & 96.7/\second{96.6}/\second{98.5} & 96.0/95.9/97.7 & 86.5/87.2/89.9 & \second{98.1}/\best{97.6}/98.3 & \best{99.9}/88.4/\best{98.8} \\
Lettered Tablet & 94.5/94.1/97.2 & 91.2/91.3/96.4 & 92.3/93.2/96.2 & 94.3/91.4/97.0 & 92.2/80.9/97.8 & 77.2/73.1/68.3 & 87.1/84.0/89.7 & 95.1/\best{94.6}/97.3 & 94.9/94.3/97.5 & 77.4/76.0/94.4 & \second{95.5}/\second{94.4}/\second{98.2} & \best{98.3}/84.8/\best{98.9} \\
Oblong Tablet & 87.5/89.7/94.5 & 77.9/81.5/92.4 & 81.6/77.5/95.2 & \second{95.4}/\best{93.9}/\second{98.1} & 85.5/76.2/94.3 & 72.3/70.5/76.0 & 93.5/89.7/93.0 & 87.3/90.5/92.6 & 93.2/91.6/97.1 & 76.1/75.8/89.1 & 92.7/\second{92.3}/97.3 & \best{98.0}/87.3/\best{98.6} \\
Pink Tablet & 97.7/97.3/98.2 & 96.4/95.5/98.8 & \second{99.0}/\best{98.5}/98.0 & 98.8/\second{98.3}/\best{99.4} & 97.9/90.5/98.1 & 94.3/91.8/92.7 & 97.4/96.5/96.9 & 98.5/97.2/98.2 & 98.3/97.5/98.3 & 95.4/94.2/98.2 & 98.9/97.7/99.1 & \best{99.7}/91.6/\second{99.3} \\
Red Tablet & 77.1/80.6/82.9 & 77.5/79.8/71.2 & 74.9/77.6/83.6 & 87.3/\second{89.6}/\best{89.7} & 87.0/79.4/\second{88.5} & 74.3/71.3/73.7 & 80.0/81.4/66.4 & 79.1/82.4/84.1 & 82.6/85.1/85.0 & 77.5/75.5/69.5 & \second{89.0}/88.8/86.4 & \best{93.2}/\best{92.7}/88.1 \\
White Tablet & 97.2/96.0/97.6 & 96.1/95.4/96.3 & 97.9/95.9/98.2 & 98.0/96.0/98.4 & 97.5/88.3/98.2 & 88.9/88.9/82.4 & \second{99.1}/\second{96.6}/\best{99.2} & 98.2/\second{96.6}/98.7 & 98.5/95.7/97.8 & 95.5/93.9/97.2 & 98.7/\best{97.2}/98.4 & \best{99.2}/89.8/\second{99.0} \\
Yellow Tablet & 98.6/\second{98.3}/97.7 & 98.7/96.3/98.5 & 98.7/95.5/98.3 & 98.4/95.0/98.6 & 97.1/91.1/98.6 & 93.8/92.7/94.5 & 99.3/97.4/98.4 & 98.9/\second{98.3}/\second{98.9} & 98.1/97.0/97.1 & 98.3/96.5/97.6 & \second{99.4}/\best{98.6}/98.8 & \best{99.9}/95.4/\best{99.4} \\
\hline
Button & 88.7/92.5/95.2 & 76.7/86.8/93.2 & 89.1/82.6/92.3 & \second{93.9}/\best{96.8}/\best{99.6} & 87.2/79.9/99.0 & 92.8/94.5/96.8 & \second{93.9}/\second{96.0}/99.4 & 90.1/93.8/\second{99.5} & 91.0/93.8/98.5 & 92.9/88.5/99.4 & 93.2/94.9/\second{99.5} & \best{94.2}/83.8/\best{99.6} \\
Gear & 96.6/95.3/99.4 & 85.1/80.4/97.1 & 96.0/94.3/99.2 & \second{98.1}/\best{97.0}/\best{99.7} & 90.2/78.1/97.2 & 87.7/81.6/76.3 & 98.0/\second{96.9}/99.0 & 96.8/96.1/97.4 & 96.8/96.2/98.4 & 84.8/78.2/86.3 & 97.3/96.3/99.4 & \best{98.9}/92.0/\second{99.6} \\
Nut & 93.0/87.5/96.4 & 84.2/88.4/97.9 & \best{96.6}/92.2/97.9 & 94.0/93.2/97.5 & 82.0/71.2/97.2 & 66.5/80.2/86.8 & 82.1/80.4/94.4 & 91.0/\second{94.0}/99.0 & 94.1/92.8/97.8 & 92.7/89.6/97.6 & \second{96.2}/\best{96.2}/\second{99.1} & 96.1/77.7/\best{99.2} \\
Nut Cap & 94.3/89.0/97.2 & 86.2/80.3/95.0 & 92.2/87.2/96.5 & \best{98.8}/\best{92.9}/\second{98.4} & 88.8/78.9/96.1 & 75.0/65.3/55.7 & 87.3/77.1/93.0 & 93.0/86.3/96.1 & 93.8/\second{89.8}/97.6 & 89.4/79.7/91.4 & \second{94.4}/88.8/97.8 & 90.4/82.5/\best{99.2} \\
Red Washer & 97.5/97.2/98.3 & 93.0/95.0/98.6 & \second{99.1}/\second{97.6}/\second{99.2} & 97.7/96.6/98.6 & 95.4/88.1/98.4 & 96.7/94.7/91.5 & 96.7/95.8/97.7 & 96.6/95.9/98.2 & 96.9/95.0/98.4 & 94.3/91.9/98.8 & 98.5/\best{98.6}/\best{99.3} & \best{99.5}/90.1/\second{99.2} \\
Round Button* & 97.3/97.9/98.4 & 89.9/93.0/92.8 & 97.2/98.0/\second{99.5} & 95.7/96.8/98.3 & 94.6/83.2/97.9 & 82.2/82.4/85.2 & \second{99.2}/98.5/99.2 & 97.9/\second{98.6}/\second{99.5} & 98.9/98.1/\best{99.6} & 88.1/87.9/96.7 & 98.8/\best{98.9}/\second{99.5} & \best{99.9}/97.1/\best{99.6} \\
Screw & 90.8/88.8/94.6 & 71.1/76.6/91.8 & 96.6/89.0/96.7 & \second{97.7}/\second{92.8}/96.4 & 91.0/78.0/95.5 & 65.7/67.1/81.7 & 90.9/88.4/96.3 & 90.6/89.0/96.4 & 94.1/91.7/95.6 & 74.5/72.2/80.6 & 93.7/\best{94.8}/\second{99.0} & \best{99.0}/89.6/\best{99.3} \\
Square Button* & 98.2/\second{98.4}/\second{99.4} & 92.3/93.7/98.4 & 97.7/97.1/99.3 & 95.0/95.0/98.3 & 92.9/85.7/96.4 & 97.0/91.9/85.4 & 97.2/96.9/98.7 & 97.6/97.2/99.3 & 97.0/96.9/99.3 & 92.2/90.3/97.9 & \second{99.6}/\best{98.5}/99.0 & \best{99.8}/94.3/\best{99.6} \\
Terminal & 93.6/94.0/96.1 & 84.1/84.5/96.5 & 97.1/\best{98.2}/98.9 & 96.8/\second{97.6}/98.6 & 96.8/93.6/97.7 & 82.2/77.3/70.7 & 96.7/95.0/98.5 & 94.0/92.8/97.2 & 96.5/95.9/97.4 & 77.0/73.7/90.7 & \second{97.7}/97.1/\second{99.3} & \best{99.3}/90.8/\best{99.5} \\
Wire Cap & 90.6/93.0/95.0 & 83.7/84.8/94.0 & \second{96.6}/\best{95.4}/98.3 & 95.1/94.2/98.7 & 93.1/85.1/94.6 & 93.3/87.9/88.4 & \second{96.6}/\best{95.4}/98.3 & 91.0/91.1/98.5 & 94.6/93.9/96.6 & 82.6/80.8/95.6 & 93.7/\second{94.8}/\second{99.0} & \best{98.1}/88.2/\best{99.4} \\
Yellow Washer* & 92.1/91.2/95.5 & 87.5/88.3/92.8 & 96.9/91.7/95.2 & \best{97.1}/\best{96.1}/\best{96.5} & 93.1/85.1/94.6 & 78.5/76.5/77.9 & 92.8/91.1/94.0 & 93.3/\second{95.1}/95.3 & 93.4/92.4/95.1 & 89.6/85.7/90.4 & 95.0/93.6/95.9 & \second{97.0}/93.1/\second{96.2} \\
\hline
Coffee Beans & 74.9/75.5/89.5 & 74.8/72.3/84.5 & 67.7/64.3/84.4 & \second{88.9}/\second{86.8}/\second{91.5} & 70.0/64.9/82.5 & 65.1/60.4/62.3 & 83.6/80.9/88.4 & 64.4/62.1/85.5 & 85.3/80.9/89.6 & 67.3/61.5/70.3 & 73.0/71.2/83.8 & \best{89.8}/\best{87.0}/\best{91.6} \\
Goji Berries & 89.0/85.0/91.2 & 87.8/83.7/93.5 & 92.2/81.3/90.4 & \best{94.2}/\best{89.5}/\second{95.3} & 86.5/78.1/93.5 & 78.4/77.2/81.9 & 86.4/79.7/87.7 & 89.2/84.3/93.1 & 92.4/\second{87.5}/93.7 & 77.7/75.1/88.6 & 90.6/87.3/94.9 & \second{92.5}/80.7/\best{95.4} \\
Pistachios & 78.1/\second{77.3}/83.6 & 74.7/75.9/77.8 & 71.7/67.0/78.4 & \best{79.8}/\best{78.3}/\best{88.1} & 76.6/69.8/82.5 & 57.2/62.8/67.5 & 61.2/60.8/69.1 & 74.6/72.9/81.5 & 78.4/77.0/\second{86.7} & \second{79.4}/76.6/76.7 & 75.4/75.0/81.3 & 75.4/73.8/81.4 \\
\hline
Average & 92.4/92.2/95.4 & 87.2/87.7/93.1 & 88.8/88.2/94.7 & \second{94.5}/\best{93.9}/\second{96.8} & 90.2/81.9/94.1 & 80.0/78.4/78.7 & 92.0/90.3/92.7 & 92.0/91.9/95.4 & 93.4/92.8/95.5 & 84.5/82.3/90.4 & 94.0/\second{93.4}/96.2 & \best{96.8}/89.6/\best{97.2} \\
\hline
\end{tabular}%
}
\end{table*}

\section{Qualitative visualization}
\begin{figure*}[ht]
\centering
\includegraphics[width=0.9\textwidth]{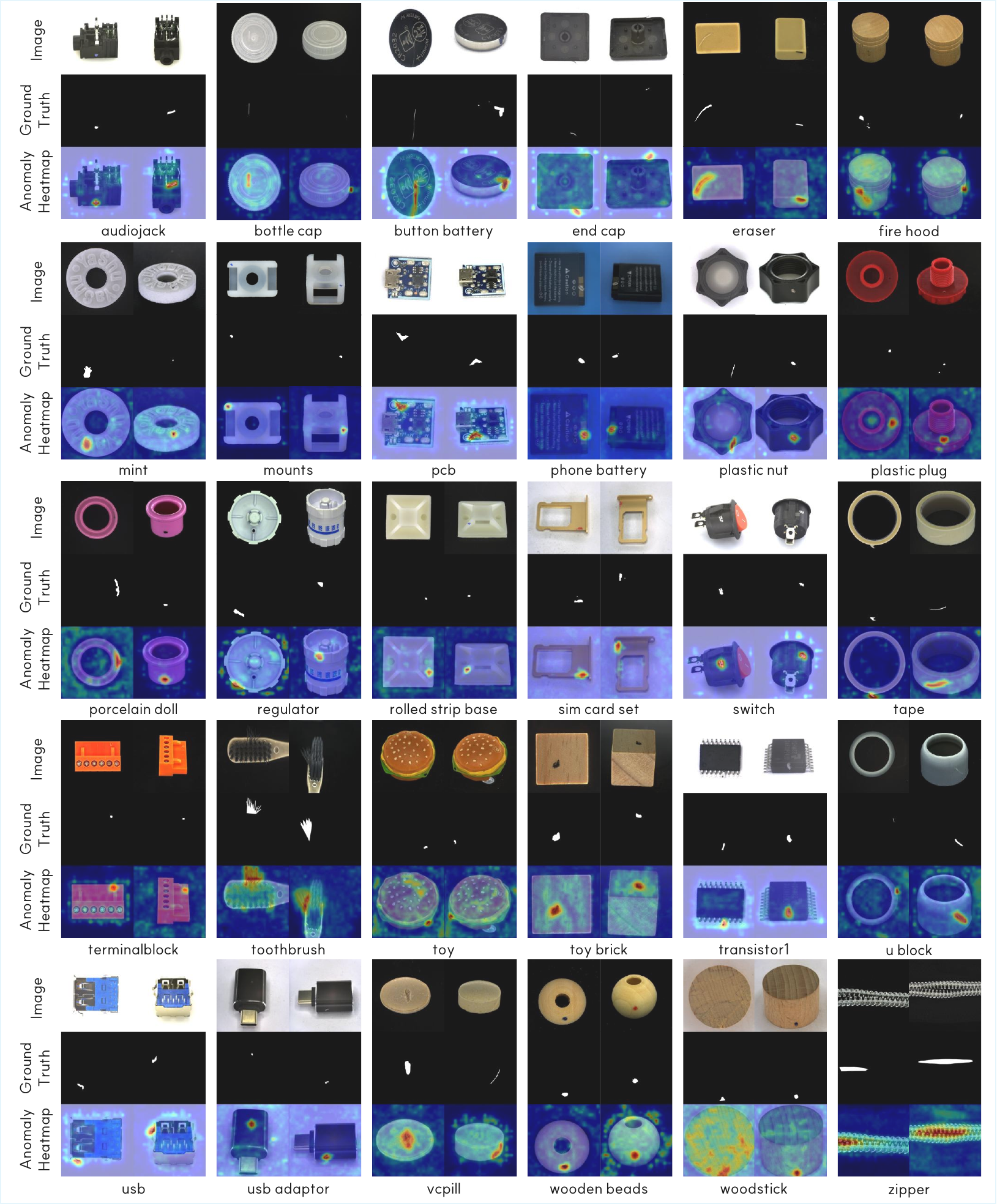} 
\caption{
     Visualization of anomaly detection results on the Real-IAD dataset.
}
\label{fig:qualitative_results_all}
\end{figure*}

We visualize the anomaly maps produced by PAR on Real-IAD in Fig.~\ref{fig:qualitative_results_all}. Compared with the ground-truth masks, our model successfully detects various defect patterns and sizes.

\end{document}